\documentclass[journal]{IEEEtran}
\usepackage{graphicx}
\graphicspath{ {./Figures/} }
\usepackage{subcaption}
\usepackage[T1]{fontenc}
\usepackage{amsmath}  
\usepackage{amssymb}
\usepackage{amsthm}
\usepackage{algorithmic}
\usepackage[procnumbered,ruled,linesnumbered,vlined]{algorithm2e} %
\SetKwInput{KwInput}{Inputs}
\SetKwInput{KwOutput}{Output}
\usepackage{soul}
\usepackage{xcolor}
\usepackage{import}
\usepackage{tkz-graph}
\usepackage{booktabs}
\usepackage{multicol}
\usepackage{adjustbox}
\usetikzlibrary{arrows}
\usetikzlibrary{positioning}
\usepackage{tikz}

\soulregister\cite7
\soulregister\ref7
\soulregister\cref7
\soulregister\autoref7

\ifCLASSINFOpdf

\else

\fi

\begin{document}
%
\title{Efficient Energy-Optimal Path Planning for Electric Vehicles Considering Vehicle Dynamics}
%
%
%

\author{Saman~Ahmadi*,~
        Guido~Tack,~
        Daniel~Harabor,
        Philip~Kilby,
        and~Mahdi~Jalili
\thanks{S. Ahmadi, G. Tack and D. Harabor are with the Department of Data Science and AI, Monash University, Australia and Mahdi Jalili is with RMIT University, Australia. SA is funded by Australian Research Council through project IE250100126, and MJ through projects DP240100963, IC240100010, IM240100042 and LP230100439. This work is also partially supported by Department of Climate Change, Energy, the Environment and Water through project number ICIRN000077.}
\thanks{* saman.ahmadi at monash.edu}
}

\maketitle

\begin{abstract}
The rapid adoption of electric vehicles (EVs) in modern transport systems has made energy-aware routing a critical task in their successful integration, especially within large-scale transport networks. 
In cases where an EV's remaining energy is limited and charging locations are not easily accessible, some destinations may only be reachable through an energy-optimal path: a route that consumes less energy than all other alternatives. 
The feasibility of such energy-efficient paths depends heavily on the accuracy of the energy model used for planning, and thus failing to account for vehicle dynamics can lead to inaccurate energy estimates, rendering some planned routes infeasible in reality.
This paper explores the impact of vehicle dynamics on energy-optimal path planning for EVs. We first investigate how energy model accuracy influences energy-optimal pathfinding and, consequently, feasibility of planned trips, using a novel data-driven model that incorporates key vehicle dynamics parameters into energy calculations.
Additionally, we introduce two novel online reweighting and energy heuristic functions that accelerate path planning with negative energy costs arise due to regenerative braking, making our approach well-suited for real-time applications.
Extensive experiments on real-world transport networks demonstrate that our method significantly improves both the computational efficiency of energy-optimal pathfinding for EVs.
\end{abstract}


%
\IEEEpeerreviewmaketitle

\section{Introduction}
\label{sec:intro}
%
%
%
%
\IEEEPARstart{T}{he} rapid advancements in automotive and renewable energy technologies over the past decade have driven the widespread adoption of electric vehicles (EVs), solidifying their role in modern transport systems. 
Today, EVs are integrated into nearly all forms of transport, serving diverse purposes such as public transit with electric buses \cite{Liu23}, daily commute with battery-powered public bikes \cite{Zhang22}, and warehouse automation with smart electric robots \cite{Bolu2019robot}.
Although technically superior to conventional vehicles in many aspects, such as powertrain system efficiency, EVs face some challenges related to their battery capacity, slow charging process and unavailability of charging infrastructure, all of which hinder their large-scale adoption. 
In scenarios where battery State of Charge (SoC) is low and charging locations are not in proximity, distant locations may only be reachable via an \textit{energy-optimum path}, meaning a path from origin to destination that minimizes energy consumption compared to all other possible paths.
It is therefore crucial to have planning algorithms that can reason about energy-optimum paths, either to calculate them directly such as in an application for satellite navigation, or as a sub-task in a bigger system such as a tool for complex vehicle routing.
In transport networks, energy-optimal pathfinding is a practical real-world application of the classic shortest path problem, aiming to find the least-cost path between two locations in a network.

The quality of the energy-efficient paths critically depends on the accuracy of the energy model used to estimate the energy requirement of each road segment (link) in the underlying network. 
In the case of EVs, energy models must account for factors such as terrain, speed, traffic, and load to provide realistic and reliable energy predictions.
While each type of EV exhibits distinct energy consumption characteristics \cite{Yi2018}, existing research often relies on energy models built based on generic vehicle categories (e.g., vehicle categories in HBEFA\footnote{Handbook of Emission Factors for Road Transport, \texttt{www.hbefa.net}}) \cite{baum2013energy,DBLP:journals/transci/BaumDGWZ19}, or simplified assumptions that do not account for system dynamics, such as focusing solely on average energy efficiency and road gradients \cite{storandt2012quick,eisner2011optimal,artmeier2010shortest}, or assuming constant speed across road segments \cite{SachenbacherLAH11,Cela0214,shen2019energy}.
A main drawback with such simplified models is that they do not fully take vehicle dynamics into account, and thus the projected energy requirements may not reflect the actual energy required to traverse specific links in the road network. 
As we will see shortly, path planning with basic energy models can produce infeasible paths, where the vehicle exhausts its available energy before reaching the designated target.

\begin{figure*}[t]
\centering
\footnotesize
\includegraphics[width=1\textwidth]{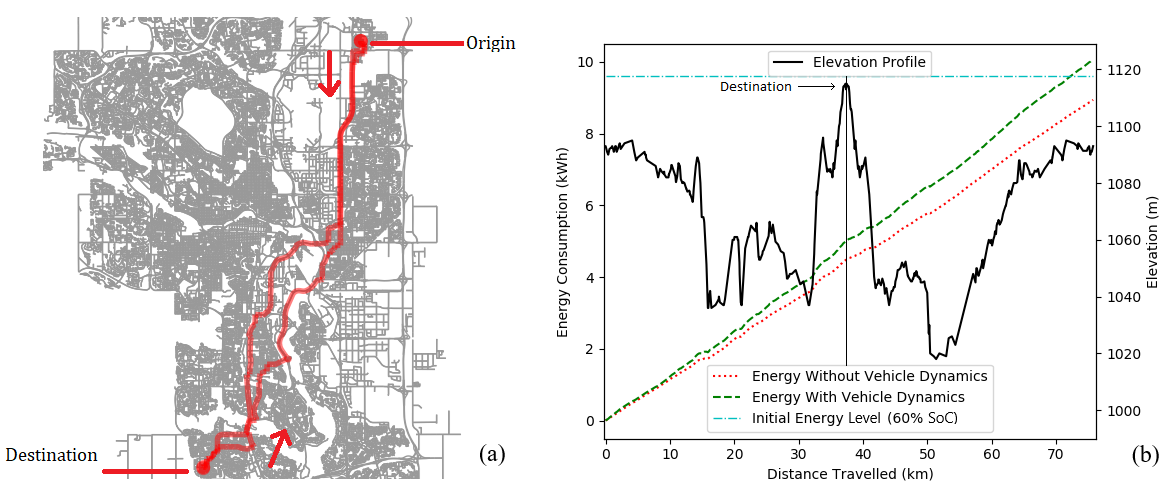} 
\caption{\small (a) Round-trip path for a sample energy-efficient route in Calgary, with arrows indicating the travel direction during the outbound segment. (b) Energy consumption and elevation profile of the same path using two energy models. With an initial SoC of 60\%, the more accurate model deems the trip infeasible, since the total energy requirement exceeds the available energy.}
\label{feasibity}
\end{figure*}
\subsection{Case Study and Motivation}
To illustrate the importance of vehicle dynamics, consider an example in Calgary, Canada, where the energy-optimal path for a round trip is planned using a basic energy model for an EV with the specifications of the \textit{Peugeot iOn} (a sample vehicle with a 16 kWh battery capacity).
We suppose there are four passengers on board and 60\% initial SoC. 
Figure~\ref{feasibity} shows the planned round-trip on the right, and changes in elevation and the aggregated energy consumption on the left.
In this figure, we compare a conventional (basic) energy model which considers average energy consumption adjusted with road gradients (dashed red curve) with a more accurate energy model which adds vehicle load and driving patterns to the basic model (dashed green curve).
The difference between the two models is 1kWh or 6.3\% of the battery capacity at the end of the trip, which is clearly not negligible. 
More importantly, the path computed with the simpler model is most likely infeasible in reality, leaving the vehicle and its passengers stranded. 
As we will study in the experiment section, such cases are not pathological: depending on the SoC, vehicle dynamics and the length of the trip, such infeasible paths can arise in a majority of cases.
We will also show that a well-designed and realistic energy model can not only improve the quality of energy estimates, but reduce the risk of producing infeasible routes, and can even yield improvements in algorithmic efficiency.

\subsection{Energy-optimal Pathfinding}
The energy-optimal pathfinding problem aims to minimize energy consumption while ensuring route feasibility, constrained by the limited available energy.
For conventional vehicles, where energy (fuel) consumption is non-negative, the problem can be efficiently solved using dynamic programming solutions such as Dijkstra's algorithm \cite{dijkstra1959note}.
Nonetheless, many EVs feature energy recuperation via braking, which may result in negative energy costs on some negative slopes. 
Baseline approaches, such as the Bellman-Ford algorithm \cite{bellman1958routing,ford1956network} are capable of solving pathfinding problems on graphs with negative weights, as in~\cite{Cedric2020,Morlock2020}. 
However, their computational complexity is higher compared to algorithms designed for non-negative cost graphs. 
More recent approaches leverage the reweighting technique of Johnson’s algorithm \cite{johnson1977efficient} to enable more efficient pathfinding, provided the graph contains no negative cycles, a condition that is inherently satisfied in our problem due to the law of conservation of energy.
%
In real world applications, however, we may face frequent changes either in the graphs (such as changes in the driving pattern in different traffic conditions) or in the vehicle specification (such as changes in energy consumption via varying load or number of passengers).
As we will see, the overall system complexity means that without a suitable energy model and pathfinding algorithm, planned routes may indeed be infeasible in practice.
Therefore, it is crucial to have path planning algorithms that are fast and accurate in the dynamic setting.

The key contributions of this article are as follows:
i) A data-driven energy model for EVs that accurately estimates the energy required for each road segment by incorporating key factors such as vehicle load, road gradients, and driving patterns.
ii) A comprehensive analysis of the impacts of vehicle dynamics on feasible, energy-optimal path planning for EVs in various settings.
iii) Yea novel reduced cost functions that allow for pre-processing-free computation of energy-optimal paths by identifying a natural characteristic of the system.

The remainder of this article is structured as follows. 
In the next section, we formally define the problem and review its classical solution approaches.
In section~\ref{sec:vehicle_dynamics}, we discuss vehicle dynamics and the key parameters involved in estimating EV energy consumption. 
Our data-driven model is presented in this section.
Section~\ref{sec:reduced_cost} focuses on energy-optimal pathfinding and demonstrates how our reweighing functions integrate with the core problem. 
Finally, in Section~\ref{sec:experiment}, we present a comprehensive analysis of the impact of vehicle dynamics on reliable pathfinding, along with an experimental evaluation that highlights the effectiveness of our proposed model in achieving faster and more accurate pathfinding for EVs.
\section{Problem Formulation and Background}
\label{sec:problem_formulation}
Let us consider a road network modeled as a directed graph $G=(V,E)$ with a finite set of vertices~$V$ and a set of edges~$E \subseteq V \times V$.
Each vertex $u \in V$ corresponds to a location (intersection) and each edge $e \in E$ corresponds to a link (road segment).
A path is then defined as a sequence of vertices $u_i \in V$ with $ i \in \{1, \dots, n \}$ such that $(u_i,u_{i+1}) \in E$ for $ i \in \{1, \dots, n-1 \}$.
In our notation, expanding a typical vertex $u$ generates a set of successor vertices, each denoted $\mathit{Succ}(u)$ and we always have $(u,\mathit{Succ}(u)) \in E$.
Thus, we always have $(u, \mathit{Succ}(u)) \in E$.
Each edge is associated with a real-valued 
weight $\mathit{cost}_e \in \mathbb{R}$, which represents the amount of energy required to traverse the link. 
The edge costs can be retrieved via the function $\mathit{cost}: E \rightarrow \mathbb{R}$.
For realistic path planning, these costs need to accurately reflect the actual energy requirement of the vehicle over the edge, and can be negative due to energy recuperation.

Assume an EV with an initial energy level of $\mathcal{E}_{init}\in\mathbb{R}^+$ at the origin $\mathit{u_o} \in V$ and with the maximum energy level of $\mathcal{E}_{max}\in\mathbb{R}^+$ (the battery capacity). 
Our objective is to find an energy minimum path $\pi^*$ from $u_o$ to destination $u_d \in V$ subject to the constraint that all locations on $\pi^*$ are reached with a valid energy level in the $[0,\mathcal{E}_{max}]$ range.

%
\subsection{Energy Constraints and Adjustments}
If standard pathfinding algorithms are used to compute minimum-energy paths without considering energy constraints, the resulting paths may consume more energy than the vehicle's battery capacity.
Let $\mathcal{E}(u) \in \mathbb{R}^+$ be the (optimal) energy level of the EV at vertex $u$.
To ensure $u$'s expansion yields valid energy levels, the energy requirement of each edge $e:(u,v)$ should be checked against two conditions during the expansion \cite{artmeier2010shortest}. 
In the first case, given $\mathit{cost}_e$ as the link's energy requirement, vertex $v$ can be reached if the EV's energy level at vertex $u$ is at least $\mathit{cost}_e$, i.e., $\mathcal{E}(u) \ge \mathit{cost}_e$.
This relation ensures feasibility via $\mathcal{E}(v) \ge 0$ where $\mathcal{E}(v)$ is the energy level at the end point $v$ after traversing the edge $(u,v)$. 
The second case deals with negative cost links, and check the energy level at the end point $v$ against the battery capacity \textit{$\mathcal{E}_{max}$} to avoid overcharging (in the planner). 
If $\mathcal{E}(v)> \mathcal{E}_{max}$, the corresponding cost needs to be (temporarily) adjusted to limit the $v$'s energy level at $\mathcal{E}_{max}$, i.e., we set $\mathit{cost}_e \leftarrow \mathcal{E}(u) - \mathcal{E}_{max}$ to have $\mathcal{E}(v)=\mathcal{E}_{max}$. This ensures the energy level in the planner is always capped at the maximum possible level $\mathcal{E}_{max}$.

\subsection{Negative Weights and Graph Reweighting}
In cases where the EV's energy model is not known, or so complex that it cannot be represented as a generic function, we can preprocess edge costs and convert them into non-negative values.
Johnson's algorithm \mbox{\cite{johnson1977efficient}} is one such shifting method that can solve the all-pairs shortest path problem in graphs with negative costs but without negative cycles.
A negative weight cycle is a cycle in which the sum of the edge costs is negative.
In our application, having negative cycles means that there exists at least one cycle in our road network in which our EVs can run forever, which is not possible given the law of conservation of energy.
Note that we assume no external charging happens while the EV traverses the links. 
Therefore, our underlying graphs contain no negative cycle, and thus the Johnson's technique \mbox{\cite{johnson1977efficient}} can be applied.

The main idea in the Johnson's method is establishing a shifting function $\mu: V \rightarrow \mathbb{R}$ that satisfies $\mathit{cost}(u,v) + \mu(u) - \mu(v) \geq 0$ for a typical edge $(u,v) \in E$ while guaranteeing the optimality of all shortest paths when using shifted weights.
It can be shown that a shifting function $\mu$ exists if and only if the graph has no negative cycle \cite{johnson1977efficient}.
One such shifting function $\mu$ can be obtained by using the Bellman-Ford \cite{bellman1958routing,ford1956network} algorithm in three simple steps:
(1) add a new dummy vertex to the graph and connect it to each vertex in $G$ with a zero-cost edge;
(2) compute the shortest paths from the newly added vertex to all other vertices using the Bellman-Ford algorithm;
(3) store for every vertex $u \in V$ its optimal cost obtained during the previous step in $\mu(u)$, then remove the added vertex.
Now, with the shifting function $\mu$ established, one can reweight $\mathit{cost}(u,v)$ to a non-negative value via $\mathit{cost}(u,v) + \mu(u) - \mu(v)$, called reduced cost function, and optimise paths for their reduced costs using the Dijkstra's algorithm, as in~\cite{storandt2012quick}.

\subsection{Related Work and Existing Solutions}
Considering the state-of-the-art solutions in the literature, we can categorise the existing energy-optimal pathfinding algorithms in two approaches: 
(i) algorithms that traditionally establish a shifting function based on an attribute of vertices in the underlying graph, such as shifting techniques in~\cite{DBLP:journals/algorithmica/BaumDPSWZ20} and \cite{eisner2011optimal};
(ii) algorithms that use a shifting function directly derived from a generic cost function, such as solutions in~\cite{SachenbacherLAH11} and \cite{shen2019energy}.
The former (approach i) relies on empirical  costs but uses nodes' elevation to apply proper shifting, and the latter (approach ii) uses a basic energy model normally obtained by simplifying EVs' energy equations.
We investigate both approaches from two aspects: their applicability to dynamic settings and the reliability of planned routes.
Approach (i) is able to work with empirical energy costs or complex models, but keeping the reduced cost updated in this approach can potentially be costly in the dynamic setting as the algorithm needs to check the validity of the shifting function upon every graph change (usually in a form of preprocessing in the static mode).
Therefore, this approach can offer more reliable routes, but it may not be fast enough for dynamic applications.
On the other hand, approach (ii) can be adapted to be used in dynamic settings as any changes in the input parameters (and consequently output via the cost function) would simply be reflected in the shifted costs. 
However, the cost function is already oversimplified and does not take all the actual and realistic energy components into account.
Therefore, this approach offers fast solutions to dynamic energy-optimal pathfinding, at the risk of producing infeasible routes.

We now recapitulate the state-of-the-art shifting function of ~\cite{DBLP:journals/algorithmica/BaumDPSWZ20}, which works with energy and elevation data (approach i).
Their height-induced potential function $\mu_h: V \rightarrow \mathbb{R}$ is defined by $\mu_h=\alpha h$ where $h$ is the elevation function and $\alpha$ is a constant factor.
The value of $\alpha$ is determined in a way that all edge costs can be safely shifted to non-negative values by ensuring $\mathit{cost} (u,v)+ \mu_h(u) - \mu_h(v) \ge 0$ for every edge $(u,v) \in E$. 
This is done by a linear scan of all the edges in the network during the prepossessing phase using the following strategy. 
For every edge $e \in E$ with the energy requirement $\mathit{cost}_e$, we first set $\alpha_e \gets \mathit{cost}_e/ \Delta h$ where $\Delta h$ is the elevation difference between the link's end points.
Then, if $\Delta h>0$ we must always have $\alpha \le \alpha _e$, otherwise, 
if $\Delta h<0$ the relation $\alpha \ge \alpha _e$ must hold.
These conditions yield lower ($\underline{\alpha}$) and upper ($\overline{\alpha}$) bounds on the coefficient $\alpha$, which is non-negative in practice.
Finally, we can choose any value for $\alpha$ in the $[\underline{\alpha},\overline{\alpha}]$ range, where $\underline{\alpha}$ is the largest $\alpha_e$ among the \textit{downhill} edges and $\overline{\alpha}$ is the smallest $\alpha_e$ among the \textit{uphill} edges.

\section{Vehicle Dynamics and Energy Efficiency}
\label{sec:vehicle_dynamics}
Energy-aware pathfinding for EVs requires system models that can estimate the electric energy required to traverse each road segment. 
In this section, we develop a novel technique that can integrate the essence of an EV's longitudinal dynamic into a cost function for energy-optimal path planning.

The energy consumption of an EV can be accurately modelled based on a so-called \emph{longitudinal vehicle dynamics}, which essentially establishes a nonlinear relationship between the vehicle speed, mass, road gradient and the required power. 
Typically, the electric power drawn from the EV's energy storage system can be calculated by the following equation in a stop-and-go pattern on a non-flat road \cite{ehsani2018modern}:
\begin{equation} \label{eq:veh_dyn}
P=\frac{V}{\eta} (M g f_r cos \theta+\frac{1}{2}\rho C_D A V^2+Mg\sin \theta+M\delta\frac{dV}{dt})
\end{equation}
%
where $\eta$ is the transmission efficiency, $f_r$ is the tire rolling resistance coefficient, $\rho$ is the air density, $C_D$ is the aerodynamic drag coefficient, $A$ is the frontal area of the vehicle and $\delta$ is the coefficient of rotary elements in the vehicle.
Each of these parameters can be determined empirically for any given EV, and can be considered constant in the so-called longitudinal vehicle dynamics.
The remaining and interesting variables in the equation are vehicle speed $V$, acceleration $dV/dt$, mass $M$ and the ground slope angle $\theta$. 
The values for speed and acceleration are both highly dependent on the driving pattern. 
One may simplify the model by assuming fixed (constant) speeds \cite{SachenbacherLAH11,shen2019energy}, but, as we observed earlier in Section~\ref{sec:intro}, ignoring acceleration will result in inaccurate energy predictions in reality.
Since the driving pattern changes with the road conditions (such as traffic congestion), predicting energy flows over trips is a difficult task.
The industry-standard approach for addressing these uncertainties involves using drive cycle tests to evaluate the vehicle's energy efficiency under various driving patterns. 

\begin{figure}[t]
\includegraphics[width=1\linewidth, height=5cm]{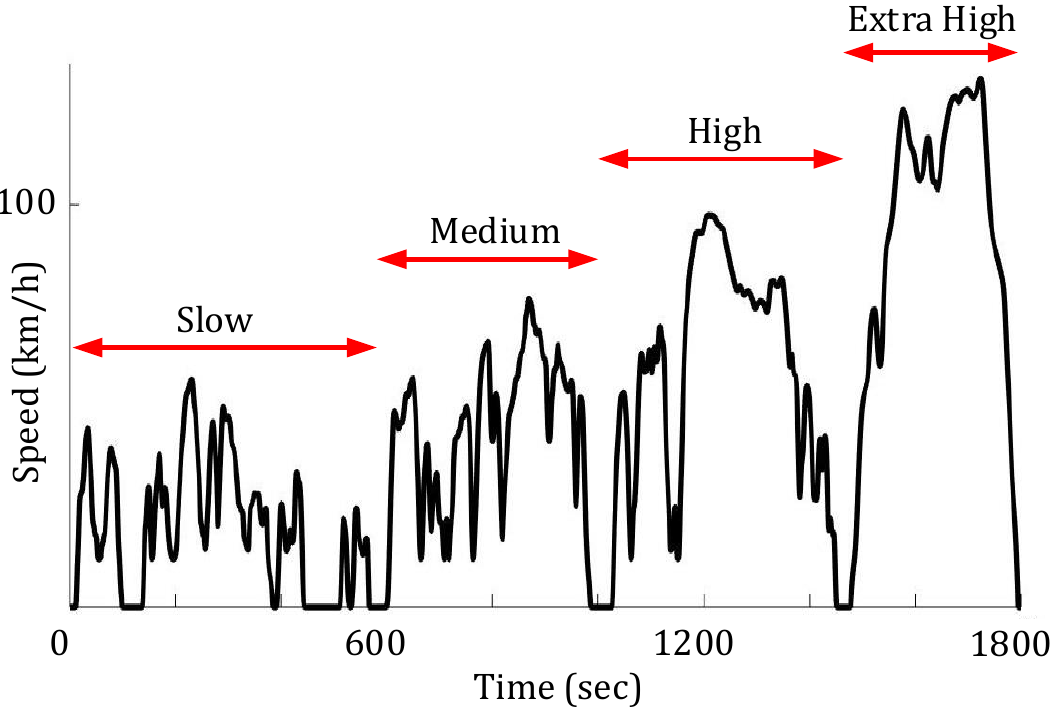} 
\caption{\small The WLTP cycle and its four speed profiles \cite{ciuffo2015development}.}
\label{wltp}
\end{figure}

While several drive cycles exist for specific scenarios, we are interested in those that encompass the most prevalent driving patterns in urban transport.
In this study, we utilize the WLTP\footnote{Worldwide Harmonised Light Vehicles Test Procedure} cycle, a realistic drive cycle developed from real-world driving data. 
Many automotive manufacturers now report their vehicles' energy efficiency based on this standardized cycle \cite{ciuffo2015development}.
The speed profile of WLTP and its speed categories (Slow, Medium, High, Extra-High) are depicted in Figure~\ref{wltp}.

\textbf{Data-driven energy model:}
While existing studies often rely on simplified longitudinal equations to estimate energy consumption, this research adopts a more robust, data-driven approach that integrates vehicle dynamics and road conditions, providing improved accuracy for real-world driving scenarios.
Specifically, we use energy data, derived either synthetically or from real-world measurements, to build a model that maps the key vehicle dynamics variables to energy efficiency. 
Given that the underlying energy requirement of EVs can be mathematically described by vehicle dynamic equations (Eq.~(\ref{eq:veh_dyn})), our experiments, as presented in Section~\ref{sec:experiment}, show that the system can be accurately represented by a second-degree polynomial with high precision (via standard regression), consistent with the quadratic form inherent in vehicle dynamics.
Consequently, the core of our approach is an energy function that measures the energy efficiency $\mathit{cost_{ef}}$ of a particular EV (usually reported in Wh/100m) under a specific driving pattern in the generic form of:
\begin{equation} \label{eq:ener_eff}
\mathit{cost_{ef}}=(ma_2+b_2)s^2+(ma_1+b_1)s+(ma_0+b_0)\\
\end{equation}
Here, $s$ is the road angle ($\sin \theta$) and $m$ is the extra mass (load/passengers' weight in kg).
The coefficients $a_i$ and $b_i$, $i \in \{0,1,2\}$, are also non-negative parameters, determined by the mathematical learning method. 
See Table~\ref{table_coefficient} in Section~\ref{sec:experiment} for the range of $a_i$ and $b_i$ parameters for studied EVs in this paper under the speed profiles of WLTP.
For synthetic data, any mathematical formulation or powertrain simulator can be employed to produce the necessary energy data for learning the energy model, provided that vehicle mass, road gradient and driving pattern can be adjusted. 
Our approach is versatile, allowing the model to be computed for different driving patterns.
This means that the coefficients of the quadratic function above can be obtained under any driving patterns, such as the speed profiles in the WLTP cycle.

\begin{figure}[!ht]
\includegraphics[width=1\linewidth, height=5cm]{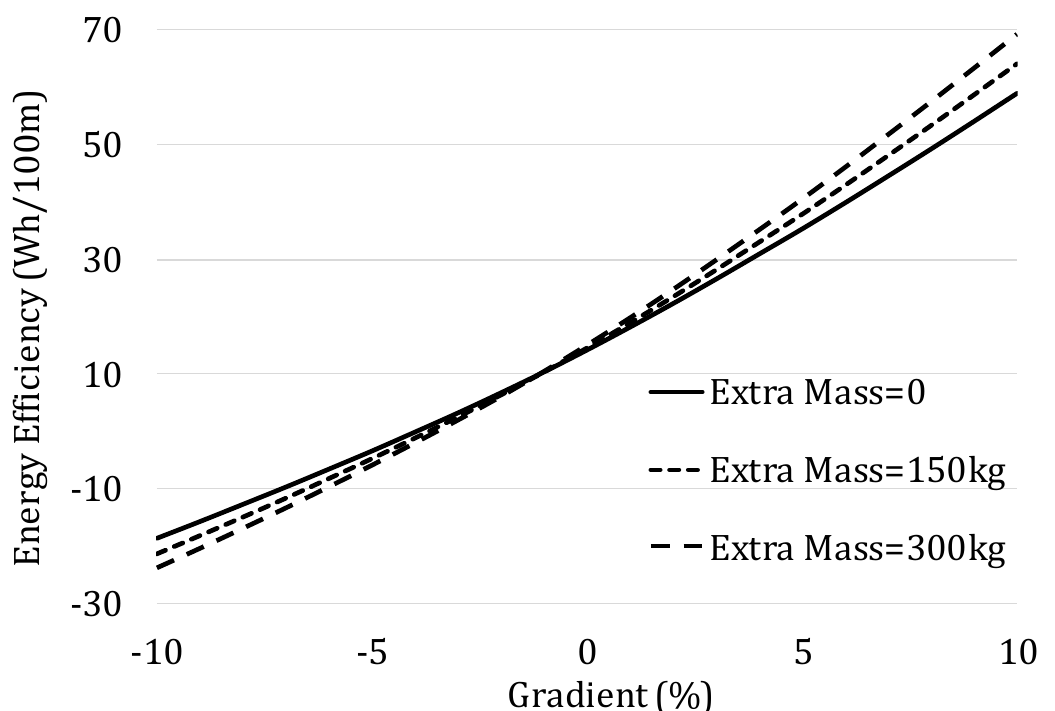}
\caption{\small Average energy efficiency for the \textit{Nissan Leaf} over WLTP in different scenarios using synthetic data.}
\label{relationship}
\end{figure}

Figure~\ref{relationship} depicts part of our simulation results for one of our test EVs, the \textit{Nissan Leaf}, with synthetic data. 
The figure shows energy efficiency in the (full) WLTP cycle with different loads and road gradients, where the initial SoC is 70\%. 
We observe a slight yet distinct non-linear relationship between road gradient and energy efficiency. 
The figure also shows that the energy an EV can recuperate on a downhill slope is significantly lower than the energy required to ascend the same gradient. This imbalance is primarily due to the braking strategy and the overall system efficiency.

%

\section{pathfinding with online reduced costs}
\label{sec:reduced_cost}
This section introduces two novel online reweighting and heuristic functions that allow standard energy-optimal pathfinding algorithms to establish a valid shifting function without graph reformulation.
We first present our model-based approach, followed by a model-independent method.
\subsection{Model-based Reduced Cost}
Given the energy model of the EV provided in the form of Eq.~(\ref{eq:ener_eff}), we can calculate the energy cost of edge~$e\in E$ as:
\begin{equation} \label{eq:ener_link}
 cost(e) = m(a_2s^2+a_1s+a_0)l + (b_2s^2+b_1s+b_0)l  
\end{equation}
where $l$ is the length of the edge $e$ (in units of 100m).
We can then develop a formula for energy consumption along path $\pi$ with $n$ edges. 
Each edge $j \in \{1,\dots,n\}$ is characterized by its length $l_j$ and road elevation angle $s_j$. 
The energy consumption along the path $\pi$ can thus be formulated as
\begin{equation} \label{eq:ener_path1}
 \mathit{cost}(\pi) =m\sum_{j=1}^{n}\sum_{i=0}^{2}(a_{i,j}s_j^i)l_j + \sum_{j=1}^{n}\sum_{i=0}^{2}(b_{i,j}s_j^i)l_j  
\end{equation}
Since each link of the path may be traversed with a different driving pattern, the parameters $a_i$ and $b_i$ are now also indexed by $j$. 
%
We observe that, since the cost function is naturally convex and all the coefficients are non-negative, the only terms that can produce negative costs are $a_{1,j}s_j$ and $b_{1,j}s_j$ for the links with negative ground slope, i.e., $s_j < 0$. 
Thus, we can define the potentially negative component of the cost as:
\begin{equation} \label{eq:ener_path2}
\mathit{cost_{neg}}(\pi) = m\sum_{j=1}^{n}a_{1,j}s_j l_j+\sum_{j=1}^{n}b_{1,j}s_j l_j 
\end{equation}
Let us analyze this potentially negative component further.
In~\cite{eisner2011optimal}, the authors show that when the underlying system is known, the reduced cost function can be obtained using the elevation difference if the energy function is linear. 
Having a nonlinear and more realistic system model, we extend that idea to find a suitable reduced cost function that fits our elevation-dependent energy model, while taking driving patterns into account.
Our first observation is that $ s_j=\sin \theta_j=\Delta h_j/l_j$, with $\Delta h_j$ denoting the elevation difference over link $j$. 
Additionally, we can approximate the $a_1$ and $b_1$ coefficients using the average values of $\overline{a_1}$ and $\overline{b_1}$ (computed over patterns) to simplify the potentially negative portion of the path cost to
\begin{equation} \label{eq:ener_pi}
cost_{neg}(\pi) \approx  
\sum_{j=1}^{n}(m\overline{a_1} + \overline{b_1})\Delta h_j
= (m\overline{a_1} + \overline{b_1})\Delta H
\end{equation}
where $\Delta H=\sum_{j=1}^n\Delta h_j$.
We can observe that the approximated energy here only depends on the elevation difference between the end points of the path, that is, $\Delta H$.
Hence, we refer to this term as the \textit{path-independent} component of the energy cost or $\mathit{cost}_{pi}$, which can be computed for any pair of locations as a deterministic cost.
Based on this important observation, we can remove the main part of possibly negative energy costs and obtain the reduced term as $cost_{\mathit{red}} = cost-cost_{pi}$.
For the range of energy coefficients and link gradients in road networks, we usually have $\mathit{cost} \ge \mathit{cost}_{pi}$ which yields non-negative reduced costs.
In exceptional cases with $\mathit{cost}<\mathit{cost}_{pi}$, we may overestimate the energy cost of the link by up to $|\mathit{cost} - \mathit{cost}_{pi}|$ to keep the reduced cost non-negative.
Our preliminary investigation shows that this situation is not easily generated in practice as the $a_1$ and $b_1$ coefficients across driving patterns are quite similar with a small deviation. 
This means that instead of link-specific values $a_{1,j}$ and $b_{1,j}$ for each vehicle type, we can use their average without losing much accuracy in our reduced cost function.
We will investigate this approximation error (measuring cases with $\mathit{cost}<\mathit{cost}_{pi}$) in the experimental section.

\textbf{Model-based energy heuristic:}
For the reduced energy cost above, we define an energy heuristic function as
\begin{equation} \label{eq:ener_heur_pi}
h_e=\delta (m{a_0}_{min}+{b_0}_{min})h_d
\end{equation}
where $h_d:V \rightarrow \mathbb{R}^+$ is an consistent distance heuristic \cite{hart1968formal}, ${a_0}_{min},{b_0}_{min}$ are the minimum of the energy coefficients $a_0$ and $b_0$ among the driving patterns, and $\delta$ is a non-negative scaling factor chosen so that for every link $(u,v)$ in $G$, we have $0 \le \delta \le \mathit{cost}_{\mathit{red}}(u,v)/ \mathit{cost}_{min}(u,v)$.
The upper bound on $\delta$ is simply obtained by a linear scan of all the edges in $G$ where $\mathit{cost}_{min}(u,v)$ is the minimum energy requirement of the ($u,v$) link (with the actual distance $l_{uv}$) defined by
\begin{equation} \label{eq:ener_min}
 cost_{min}(u,v)=(m{a_0}_{min}+{b_0}_{min})l_{uv}
\end{equation}

\subsection{Model-independent Reduced Cost}
When detailed energy data is unavailable to derive an energy model in the form of Eq.~(\ref{eq:ener_eff}), the shifting approach of~\cite{DBLP:journals/algorithmica/BaumDPSWZ20} can still be applied to reweight the graph.
This shifting function requires $\mathcal{O}(E)$ time to compute an appropriate $\alpha$-value.
This may be prohibitive when the network changes dynamically (e.g. when the EV specification changes), which would require $\alpha$ to be re-calculated.
To eliminate this overhead, we borrow a basic law from physics and present our model-independent reduced cost as $\mathit{cost}_{\mathit{red}} = \mathit{cost} - \mathit{cost}_{\mathit{pot}}$ with $\mathit{cost}_{\mathit{pot}}$ denoting the change in the gravitational potential energy obtained by its classic definition in physics as
\begin{equation} \label{eq:ener_pot}
cost_{pot}= (M+m)g\Delta H
\end{equation}
where $M+m$ is the total mass of the EV (vehicle kerb mass $M$ and additional load $m$), $g$ is the acceleration of gravity (9.8 N/kg) and $\Delta H$ represents the height difference.
Therefore, this reduced cost can be utilized in online settings, as its required parameters (elevation data and vehicle mass) are generally known or can be reasonably estimated.
Since the energy requirement for each link is calculated independently, we can assert that using the potential cost function described above always results in non-negative reduced energy costs (see Appendix~\ref{sec_appendix_A} for the proof).


\textbf{Model-independent energy heuristic:}
We define our energy-based heuristic for the reduced cost above as
\begin{equation} \label{eq:ener_heur_pot}
h_e=\lambda \eta_e  h_d
\end{equation}
where $h_d : S \rightarrow \mathbb{R}^+$ is a consistent distance heuristic function, $\eta_e $ represents the average energy efficiency of the EV reported by the EV manufacturer over a standard driving cycle (with the units in Wh/100m) and the coefficient $\lambda$ is a non-negative scaling factor.
The scaling factor $\lambda$ is chosen so that for every link $(u,v)$ of the road network we have $0 \le \lambda \le \mathit{cost}_{\mathit{red}}(u,v)/(\eta_e l_{uv})$.
The upper bound on $\lambda$ is obtained by a linear scan over the edges in $G$.

\subsection{Energy-optimal A* Search with Reduced Costs}
\begin{algorithm}[t]
\footnotesize
\caption{{Energy-optimal A* Search}}
\label{alg:Dij}
\DontPrintSemicolon
 \KwInput{{A problem instance ($G$, $\mathit{cost}$, $h$, $\mathit{u_o}$, $\mathit{u_d}$, $\mathcal{E}_{\mathit{init}}$)}}
\KwOutput{Energy requirement of the optimal path}

$\mathcal{Q} \gets \emptyset$\;
$ \mathcal{P}(\mathit{u}) \gets \varnothing$, $\mathcal{C}(u) \gets \infty $, $\mathcal{E}(u) \gets -\infty $ for all $u \in V$\;
$\mathcal{C}(\mathit{u_o}) \gets 0 $, $\mathcal{E}(\mathit{u_o}) \gets \mathcal{E}_{init}$, {$f(\mathit{u_o}) \gets h(\mathit{u_o}) $ }\;

Add $\mathit{u_o}$ to $\mathcal{Q}$\;
\While{$\mathcal{Q} \neq \emptyset$}
{
{Remove from $\mathcal{Q}$ a vertex $u$ with the smallest $f$-value }\label{alg2:iter}\;
     \lIf{$u=\mathit{u_d}$}
     {{\bf break} \label{alg2:terminate}  }
        \ForEach{$ v \in Succ(u)$}
        { 
        $\mathit{cost_e} \gets \mathit{cost}(u,v)$ \;
        \lIf{{$\mathcal{E}(u) < \mathit{cost_e} + h(v)$} \label{alg2:energy-check1}}
            {{\bf continue}}
        \lIf{$\mathcal{E}(u) - \mathit{cost_e} > \mathcal{E}_\mathit{max} $}
        {$\mathit{cost_e} \gets \mathcal{E}(u) - \mathcal{E}_\mathit{max}$ \label{alg2:energy-check2}}
        
        {$ cost_{red} \gets \mathrm{Max}(0,\mathit{cost_e} - \mathit{cost_{pi/pot}}(u,v))$ \label{alg2:energy-red1}}
        
        \If{$\mathcal{C}(u) + \mathit{cost_{red}} < \mathcal{C}(v)$ \label{alg2:relax-check1}}
        {
        $\mathcal{C}(v) \gets \mathcal{C}(u) + \mathit{cost_{red}}$ \;
        {$f(v) \gets \mathcal{C}(v) + h(v)$ }\;
        $\mathcal{E}(v) \gets \mathcal{E}(u) - \mathit{cost_e}$ \;
        $\mathcal{P}(v) \gets u$ \;
        \lIf{$v \notin \mathcal{Q}$}
        {add $v$ to $\mathcal{Q}$}
        \label{alg2:relax-check2}
        }
        
        }
}
\Return{$\mathcal{E}_{\mathit{init}} - \mathcal{E}(\mathit{u_d})$}\;
\end{algorithm}
We now explain how our reduced cost functions can be incorporated into heuristic-guided pathfinding algorithms, such as A* search \cite{hart1968formal}, to identify energy-efficient paths.
We present the pseudocode of our energy-optimal A* search with energy constraints in Algorithm~\ref{alg:Dij}, which differs from the approach of \cite{SachenbacherLAH11} in both the reduced cost functions and the energy heuristics.
This algorithm can be easily converted to Dijkstra's algorithm by relaxing the heuristic function and manually setting $h(u)=0$ for every $u \in V$.
Algorithm~\ref{alg:Dij} starts with initializing the priority queue $\mathcal{Q}$, the parent array $\mathcal{P}$, and the arrays $\mathcal{C}$ and $\mathcal{E}$, which keep track of the smallest reduced cost and available energy at each vertex, respectively. 
Given $\mathcal{E}_{\mathit{init}}$ as the initial energy, the algorithm then sets the reduced cost and energy estimates at the origin location and inserts $u_o$ into the queue to commence the search.
For every vertex $u$ touched in the search, the $\mathit{start}$-$\mathit{goal}$ reduced cost estimate consists of the known cost from $u_o$ to $u$ and the estimated cost ($h$-value) from $u$ to $u_d$ namely as $f(v)=\mathcal{C}(u)+h(u)$.
Vertices in the priority queue of A* are ordered based on their reduced cost estimate (i.e., their $f$-value).
In every iteration of the search, the algorithm extracts from $\mathcal{Q}$ a vertex with the smallest $f$-value among all vertices present in the queue (line~\ref{alg2:iter}).
Let this extracted vertex be $u$.
If $u$ is the destination location $u_d$ (line \ref{alg2:terminate}), we can terminate the search and return $\mathcal{E}_{\mathit{init}}-\mathcal{E}(u_d)$ as the energy cost of the optimum path.
Otherwise, the search use $u$'s successors to extend the partial path. 
Let $v$ be one such successor vertex.
During this extension, the algorithm checks $\mathit{cost}_e + h(v)$ (estimated energy requirement to destination) against the $\mathcal{E}(u)$ (available energy at $u$) to make sure the EV has minimum (estimated) amount of energy needed to reach $u_d$ (line \ref{alg2:energy-check1}), or to check whether the link's energy needs to be adjusted (line \ref{alg2:energy-check2}).
The energy adjustment is only done when the EV's battery cannot fully capture the recuperated energy in some negative slopes.
Then, the (always non-negative) reduced energy $\mathit{cost}_{\mathit{red}}$
is calculated via Eq.~(\ref{eq:ener_pot}) or Eq.~(\ref{eq:ener_pi}) depending on the reweighting approach (line~\ref{alg2:energy-red1}).
Finally, if extending the current partial path via link $(u,v)$ reaches $v$ with a reduced cost smaller than $\mathcal{C}(v)$, we can update this reduced cost and insert $v$ into the queue for further extensions (if it is not already present in $\mathcal{Q}$).
Note that, if traversing the link $(u,v)$ with $\mathcal{E}(u)$ is feasible, the available energy at the successor vertex $v$ would be $\mathcal{E}(u)-\mathit{cost}_e$.
A complete proof of correctness of our energy-optimal A* algorithm is provided in Appendix~\ref{sec_appendix_A}.


\section{Experimental Analysis}
\label{sec:experiment}
This section studies the performance of our energy-optimal path planner with the proposed online reduced costs, and also investigates the impact of vehicle dynamics on the quality of energy-optimal paths planned for EVs.
To evaluate our contributions within a real-world transport context, we consider road networks of six different cities with various road categories: Munich, Milan, Calgary, Canberra, Vancouver, and San Francisco.
The map data was sourced from OpenStreetMap contributors~\footnote{\texttt{www.openstreetmap.org}} using the Python package OSMnx \cite{boeing2017osmnx}, focusing on the central area of each city.
We enriched each map with elevation and real travel time data using the \textit{Bing}, \textit{Mapbox}
and \textit{Here} APIs\footnote{\texttt{www.bing.com; mapbox.com; developer.here.com}}. 
Average gradients are given in Table \ref{table_Avg_Dist} while network sizes can be found in Table \ref{table:result_alg_leaf}. 

\textbf{EV simulator:}
This research utilizes the powertrain simulator ADVISOR (ADvance VehIcle SimulatOR)\footnote{\texttt{http://adv-vehicle-sim.sourceforge.net/}} to generate realistic energy data.
ADVISOR uses the MATLAB engine to simulate the vehicle's powertrain using both backward (solving dynamic equations from the driving cycle to the energy source) and forward (analysing energy flow from the energy source to wheels) simulations, based on mathematical models of all components of the EV and detailed dynamic equations such as Eq.~(\ref{eq:veh_dyn}). 
Vehicle models in ADVISOR represent actual interactions between components based on input/output relationships.
Blocks in ADVISOR include, but are not limited to, equations in vehicle dynamics, system efficiencies, thermal properties and extra features such as auxiliary electricity loads \cite{ahmadi2018improving}.
Therefore, it provides more accurate estimates on EVs' energy efficiency under realistic scenarios, such as driving a non-empty vehicle with the stop-and-go pattern on road segments with a positive/negative slope.

\begin{figure*}[!ht]
  \centering
  \footnotesize
  \includegraphics[trim=00 220 00 220, clip,width=1\linewidth,scale=0.4]{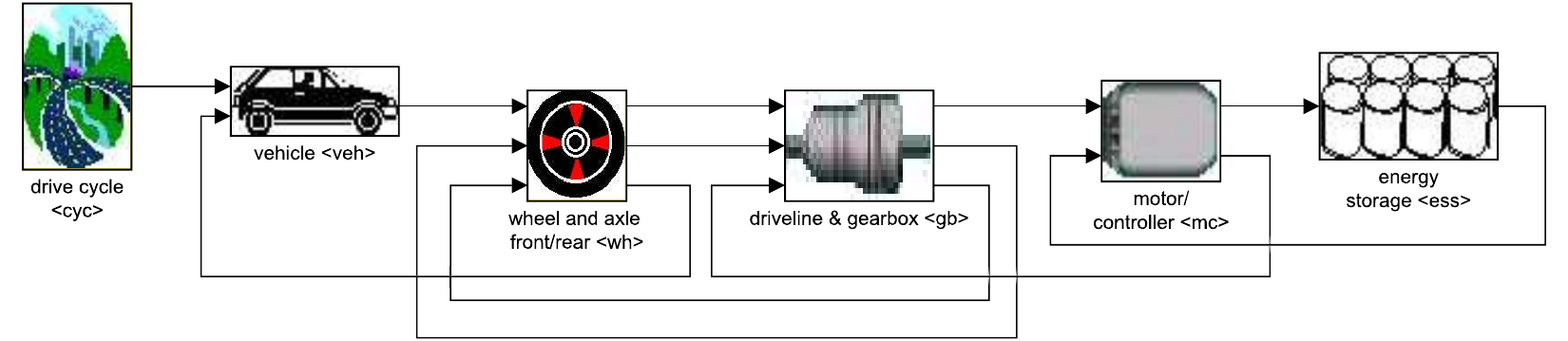}
  \caption{\small EV block diagram in ADVISOR, simultaneous backward and forward simulations}
  \label{fig:Advisor}
\end{figure*}

Figure~\ref{fig:Advisor} shows ADVISOR's high-level block diagram for EVs including the main components in the EV powertrain.
The backward simulation (left-right arrows $\rightarrow$) calculates the required energy to meet the given speed in the driving cycle, while the power delivered to the wheels is measured through the forward simulation (right-left arrows $\leftarrow$) which is more realistic when the speed is continuously changing in different traffic conditions.
More accurately, the forward simulation is a response to the backward simulation and calculates the actual amount of energy the EV's battery can supply during the cycle.

\begin{table}[t]
  \centering
  
  \footnotesize
  \caption{\small Summary of simulation parameters used for modeling energy consumption.}
  \centering
\begin{tabular}{l|l}
\hline
{Parameter} & {Variants / Range} \\
\hline
{Electric vehicle models} & \textit{Peugeot iOn}, \textit{Nissan Leaf}, \textit{GM EV1} \\
{Driving patterns (WLTP)} & Slow, Medium, High, Extra High, Full \\
{Road grade (\%)} & $[-10, -9, \dots, 9, 10]$ \\
{Additional load (kg)} & $[0, 75, 150, 225, 300]$ \\
\hline
\end{tabular}
  \label{tab:parameters}
  
\end{table}

\textbf{System models and selected EVs:}
We now explain how advanced engineering simulations with realistic driving can be utilized to extract accurate models in the generic form of Eq.~(\ref{eq:ener_eff}).
We first used the driving patterns of the WLTP cycle to generate energy efficiency data under five distinct driving patterns, 21 road gradients, and five mass profiles for three selected EVs:
the \textit{Nissan Leaf} (one of the
best-selling EVs~\cite{singer2016consumer}), the \textit{Peugeot
iOn} (used in the literature, \cite{baum2013energy,Cedric2020}) and the \textit{General Motors EV1}, a generic EV model predefined in our simulator.
Table~\ref{tab:parameters} presents a summary of the parameters used in this experiment, which resulted in over 1,500 simulation runs.
For each vehicle and driving pattern, we then applied the standard polynomial regression technique to learn an energy model in the form of Eq.~(\ref{eq:ener_eff}) using the synthesized energy data.
We obtained high precision across all models ($R^2 = 0.99$, a measure of the squared error relative to the mean error, with values close to 1 indicating an excellent fit to the training data), and the coefficients of higher-order terms were found to be nearly zero.
Table~\ref{tab:avg_coef} presents the coefficients of the energy models learned for the selected EVs of this study under the WLTP's speed profiles.

\begin{table}[t]
\footnotesize
\centering
\caption{\small EVs energy specifications and coefficients (in {Wh/100m}) over the WLTP profiles. Energy efficiency is for the entire cycle.}
{\setlength{\tabcolsep}{1.7pt}
\begin{tabular}{l|l|c|c}
\hline
Vehicle details& Profile & \textit{$[a_2,a_1,a_0]$} & \textit{$[b_2,b_1,b_0]$} \\
\hline
\textit{Nissan Leaf 2018} & Slow & [0.509, 0.238, 0.004] & [671.4, 362.9, 16.12]\\
Capacity:40kWh & Medium & [0.429, 0.241, 0.004] & [539.0, 370.4, 13.03]\\
Kerb weight:1544kg & High & [0.472, 0.249, 0.003] & [528.2, 382.8, 12.80]\\
Efficiency: & ExtraHigh & [0.829, 0.283, 0.002] & [677.9, 415.4, 15.43]\\
$\sim$14.1 $\frac{Wh}{100m}$& Overall & [0.595, 0.258, 0.003] & [602.5, 389.2, 14.24]\\
\hline
\textit{Peugeot iOn 2017} & Slow & [0.398, 0.244, 0.005] & [315.3, 264.7, 12.60]\\
Capacity:16kWh & Medium & [0.451, 0.241, 0.004] & [381.9, 262.3, 10.04]\\
Kerb weight:1050kg& High & [0.526, 0.249, 0.004] & [511.1, 259.7, 10.36]\\
Efficiency:  & ExtraHigh & [0.731, 0.262, 0.004] & [734.5, 293.1, 13.31]\\
$\sim$11.5 $\frac{Wh}{100m}$& Overall & [0.579, 0.251, 0.004] & [536.7, 272.8, 11.65]\\
\hline
\textit{GM EV1} & Slow & [0.382, 0.261, 0.005] & [505.1, 374.5, 12.44] \\
Capacity:27kWh & Medium & [0.311, 0.271, 0.004] & [325.9, 388.0, 10.43]\\
Kerb weight:1450kg & High & [0.485, 0.284, 0.003] & [354.5, 397.0, 10.46]\\
Efficiency:  & ExtraHigh & [0.632, 0.291, 0.004] & [645.7, 428.9, 12.70]\\
$\sim$11.5 $\frac{Wh}{100m}$& Overall & [1.473, 0.227, 0.002] & [608.3, 397.3, 11.25]\\
\hline
\end{tabular}}
\label{table_coefficient}
\end{table}

\textbf{Graph energy costs:}
Following~\cite{baum2013energy}, we carefully match the realistic (average) speed of every link in our road graph with one of the driving patterns in the WLTP cycle to later select a set of coefficients that best describes the energy consumption profile of our selected EV for that link. 
To this end, we first retrieve the average speed of the links through several online sources. 
Then, we compare each link’s speed with the average speed of every driving pattern and choose the closest pattern that matches the link’s average speed. 
The result is a realistic set of energy weights (in Wh/100m), provided in Table~\ref{table_coefficient}.
We can see that the coefficients $a_1$ and $b_1$ are nearly in the same range for each vehicle type.
Therefore, the function $\mathit{cost_{pi}}$ presented in Eq.~(\ref{eq:ener_pi}) is fairly accurate.
To enable our pathfinding algorithm to work on energy-weighted graphs, we use the coefficients in Table~\ref{table_coefficient}, along with elevation and speed data, to precompute edge weights for a specified load.

\subsection{Parametric Comparison of Approaches}
Table~\ref{tab:avg_coef} provides the practical range of the parameters for each pathfinding technique described in this article, including edge shifting using Johnson's algorithm $\pi_h$, shifting with potential energy $\mathit{cost_{pot}}$ and shifting with path-independent energy $\mathit{cost_{pi}}$.
The coefficients associated with $\mathit{cost_{pot}}$ and $\mathit{cost_{pi}}$ are vehicle-specific and fixed among networks, but the bounds on the coefficient $\alpha$ are calculated based on the energy requirements in road networks and vary in each graph.
As an example, the range of $\alpha$ in Table~\ref{tab:avg_coef} has been calculated for the given EVs operating in Calgary without any passenger or extra load.
A quick look over the values in this table shows that, for every vehicle type, the coefficients $Mg$ and $\overline{b_1}$ are within the bounds of $\alpha$.
Therefore, the reduced cost functions do not require any energy adjustment, and the energy estimation remains intact.
We will further investigate this in detail for different maps with varying loads.
\begin{table}[t]
  \centering
  \footnotesize
  \caption{\small Coefficients of each shifting technique (Wh/100m). Values for $\mu_h$ are for the network of Calgary without any passenger.}
  \setlength{\tabcolsep}{4.0pt}
\begin{tabular}{l| c | cc | cc | cc}
\hline
& Mass & \multicolumn{2}{c|}{$\mathit{cost}_{pot}$} & \multicolumn{2}{c|}{$\mathit{cost}_{pi}$} & \multicolumn{2}{c}{$\mu_{h}$}\\ \cline{3-8}
& & & & & & &\\[-1.4\medskipamount]
Vehicle & $kg$ & $g$ &  $Mg$ & $\overline{a_1}$ & $\overline{b_1}$ & $\underline{\alpha}$ &  $\overline{\alpha}$\\
\hline
\textit{Nissan Leaf}  & 1544kg & 0.273 & 421.51 & 0.253& 382.87 & 218.18 & 541.98\\

\textit{Peugeot iOn}  & 1050kg & 0.273 & 286.65 & 0.249 & 269.92 & 138.60 & 388.92\\

\textit{GM EV1}  & 1450kg & 0.273 & 395.85 & 0.277 & 397.09 & 275.16 & 505.42 \\
\hline
\end{tabular}
  \label{tab:avg_coef}
\end{table}

\subsection{Algorithmic Performance}
Our first set of experiments evaluates the impacts of the presented reduced cost functions on the performance of energy-optimum pathfinding.
For this experiment, our energy coefficients are characteristic for the \textit{Nissan Leaf} 
(with details shown in Table~\ref{table_coefficient}).
We assume an initial battery SoC of 70\% and with three on-board passengers\footnote{75kg per passenger based on European Directive 95/48/EC}.
All the energy weights are obtained using our realistic energy model in Eq.~(\ref{eq:ener_eff}) and we run all the algorithms on the same graph.

We compare our online reduced-cost approaches, which use Dijkstra's and A* algorithms, with the Bellman-Ford \cite{artmeier2010shortest} and Johnson (standard edge shifting with potential functions) \cite{SachenbacherLAH11} algorithms.
For the case of shifting with $\mu_h$ (the approach in \cite{DBLP:journals/algorithmica/BaumDPSWZ20}), we set $\alpha$ to be the average of its lower and upper bounds.
All algorithms were implemented in Python 3.6 using the NetworkX 2.3 package \cite{hagberg2008exploring}.
Timings are based on the average of five individual runs reported in \textit{seconds}, running on a single core of an Intel Core i7-6700 with 3.4GHz and 16GB of RAM.
Table~\ref{table:result_alg_leaf} presents the network sizes and results for around 1,000 random \textit{(origin, destination)} pairs as point-to-point trips using the \textit{Nissan Leaf}.
The results of the extended experiments for two other EVs of this study with different loads can be found in Appendix~\ref{sec_appendix_B}.
\begin{table}[t]
\centering
\footnotesize
\caption{\small Experiment results and parameters for \textit{Nissan Leaf} with three passengers. The last column shows average search expansions.}
\setlength{\tabcolsep}{3.0pt}
\begin{tabular}{l|l|l | r| *{2}{r} | r}
\hline
City & Algorithm & Algorithm &\multicolumn{3}{c|}{ Runtime(s)} & \multicolumn{1}{c}{{Exp.}}\\ \cline{4-6} 
Network & Name & Parameters &  Prep.& Avg & Max & $\times10^3$\\
\hline

Munich & Bellman-Ford &  & 0.000 & 10.13 & 16.96 & 84.2 \\
Nodes: & Johnson &   & 1.316 & 0.329 & 0.642 & 7.8\\ 
13,974 & Johnson-$\pi_h$ & $\alpha \in$[236, 616]  & 0.394 &  0.324 & 0.697 & 7.8\\
Arcs: & Dijk-$\mathit{cost_{pot}}$ & $mg+Mg$ = 483 &  0.000 &  0.317 & 0.658 & 7.8\\
36,228 & Dijk-$\mathit{cost_{pi}}$ & $m\overline{a}+\overline{b}$ = 440 & 0.000 &  0.317 & 0.635 & 7.8\\
      & {A*- $\mathit{cost}_{pot}$} & $\lambda$ = 0.893  & 0.393  & 0.162 & 0.574 & 3.6\\
      & {A*- $\mathit{cost}_{pi}$} & $\delta$ = 1.009  & 0.394  & 0.148 & 0.563 & 3.3\\
\hline

Milan & Bellman-Ford &  &  0.000 &  6.472 & 11.69  & 68.2\\
Nodes: & Johnson &  &  1.019 & 0.249 & 0.514 & 7.1\\ 
13,377 & Johnson-$\pi_h$ & $\alpha \in$[235, 616] &  0.285 &  0.244 & 0.533 & 7.1\\
Arcs: & Dijk-$\mathit{cost_{pot}}$ & ${mg+Mg}$ = 483 &  0.000 &  0.239 & 0.501 & 7.1\\
26,539 & Dijk-$\mathit{cost_{pi}}$ & $m\overline{a}+\overline{b}$ = 440 &  0.000 &  0.239 & 0.474 & 7.1\\
  & {A*- $\mathit{cost}_{pot}$} & $\lambda$ = 0.892  & 0.286   & 0.117 & 0.389 & 3.1\\
 & {A*- $\mathit{cost}_{pi}$} & $\delta$ = 1.009 & 0.285   & 0.106 & 0.349 & 2.8\\
\hline

Calgary & Bellman-Ford &  &  0.000 &  24.03 & 47.33  & 221.5\\
Nodes: & Johnson &  &  3.016  & 0.693 & 1.520 & 17.1\\ 
32,603  & Johnson-$\pi_h$ & $\alpha \in$[253, 615] &  0.802 &  0.675 & 1.467 & 17.1\\
Arcs: & Dijk-$\mathit{cost_{pot}}$ & $mg+Mg$ = 483 &  0.000 &  0.665 & 1.404 & 17.0 \\
77,172 & Dijk-$\mathit{cost_{pi}}$ & $m\overline{a}+\overline{b}$ = 440 &  0.000 &  0.665 & 1.419 & 17.1\\
 & {A*- $\mathit{cost}_{pot}$} & $\lambda$ = 0.891  & 0.801   & 0.335 & 1.099 & 7.7\\
 & {A*- $\mathit{cost}_{pi}$} & $\delta$ = 1.009 & 0.803   & 0.301 & 1.033 & 6.9\\
\hline

Canberra & Bellman-Ford &  & 0.000 &  12.63 & 20.12 & 120.7\\
Nodes: & Johnson &  & 2.646 &  0.466 & 1.022 & 11.4\\ 
23,594 & Johnson-$\pi_h$ & $\alpha \in$[253, 615] &  0.571 &  0.459 & 1.001 & 11.5\\
Arcs: & Dijk-$\mathit{cost_{pot}}$ & $mg+Mg$ = 483 &  0.000 &  0.451 & 1.027 & 11.5\\
52,112 & Dijk-$\mathit{cost_{pi}}$ & $m\overline{a}+\overline{b}$ = 440 &  0.000 &  0.451 & 0.964 & 11.5\\
  & {A*- $\mathit{cost}_{pot}$} & $\lambda$ = 0.891  & 0.570   & 0.233 & 0.944 & 5.3\\
 & {A*- $\mathit{cost}_{pi}$} & $\delta$ = 1.009  & 0.573  & 0.211 & 0.898 & 4.8\\
\hline

Vancouver & Bellman-Ford &  &  0.000 &  2.938 & 5.131  & 20.8\\
Nodes: & Johnson &   &  1.051 &  0.179 & 0.380 & 3.7\\ 
7,599 & Johnson-$\pi_h$ & $\alpha \in$[237, 629] &  0.247 &  0.176 & 0.379 & 3.7\\
Arcs: & Dijk-$\mathit{cost_{pot}}$ & $mg+Mg$ = 483 &  0.000 & 0.171 & 0.361 & 3.7\\
23,012  & Dijk-$\mathit{cost_{pi}}$ & $m\overline{a}+\overline{b}$ = 440 & 0.000 &  0.170 & 0.357 & 3.7\\
  & {A*- $\mathit{cost}_{pot}$} & $\lambda$ = 0.891  & 0.246  & 0.084 & 0.382 & 1.6\\
 & {A*- $\mathit{cost}_{pi}$} & $\delta$ = 1.009 & 0.246  & 0.077 & 0.369 & 1.5\\

\hline

San Fran.  & Bellman-Ford &  &  0.000 &  4.505 & 9.410 & 35.7\\
Nodes: & Johnson &  &  2.582 &  0.205 & 0.486 & 4.5\\ 
9,555 & Johnson-$\pi_h$ & $\alpha \in$[230, 617]  &  0.285 &  0.198 & 0.460 & 4.5\\
Arcs: & Dijk-$\mathit{cost_{pot}}$ & $mg+Mg$ = 483 &  0.000 &  0.193 & 0.474 & 4.5\\
26,809  & Dijk-$\mathit{cost_{pi}}$ & $m\overline{a}+\overline{b}$ = 440 &  0.000 &  0.193 & 0.460 & 4.5\\
 & {A*- $\mathit{cost}_{pot}$} & $\lambda$ = 0.892 &  0.285 &  0.094 & 0.444 & 1.9\\
 & {A*- $\mathit{cost}_{pi}$} & $\delta$ = 1.009 &  0.284 & 0.086 & 0.445 & 1.8\\
\hline
\end{tabular}
\label{table:result_alg_leaf}
\end{table}

Table \ref{table:result_alg_leaf} gives Bellman-Ford times as a baseline. 
This table reports the values of parameters used in each algorithm and the runtime and pre-processing time (in seconds) separately for each algorithm.
Moreover, all of the queries were solved to optimality, and we verified that all algorithms found identical solutions.
Table~\ref{table:result_alg_leaf} also shows the lower and upper bounds on the coefficient $\alpha$ in all instances.
Since the range of the graph-specific coefficient $\alpha$ in every road network of the experiment fully covers our vehicle-specific coefficients for both $\mathit{cost_{pi}}$ and $\mathit{cost_{pot}}$ functions, i.e., we have $(m+M)g \in [\underline{\alpha},\overline{\alpha}]$ and $(m\overline{a}+\overline{b}) \in [\underline{\alpha},\overline{\alpha}]$,
we can ensure that there is no error associated with our reduced costs and thus the shifting process.

According to the results in Table~\ref{table:result_alg_leaf}, the reweighting-based algorithms can significantly improve the overall run time (up to 30 times faster) compared to Bellman-Ford.
We can see that edge sifting with $\mu_h$ offers smaller pre-processing times to compute potentials compared to the traditional approach in the Johnson's technique.
Meanwhile, our implementation of Dijkstra's algorithm using $\mathit{cost_{pi}}$ and $\mathit{cost_{pot}}$ eliminates the preprocessing phase of Johnson's and presents slightly faster execution times (up to 5\%). 
It is therefore particularly useful in situations where graph structure changes (e.g., due to changing traffic conditions or number of passengers).
In such dynamic settings, online computation of reduced costs enables fast replanning of energy-optimal paths, eliminating the need for graph reweighting required by existing approaches.

Integrating consistent, admissible energy heuristics into A* significantly improves performance over Dijkstra, decreasing both runtime and number of expansions (search iterations) by up to a factor of three on average, which enables faster, more scalable route planning.
Further, the variant using reduced cost $\mathit{cost}_{pi}$ shows a slightly faster run time and fewer vertex expansions compared to A*-$\mathit{cost}_{\mathit{pot}}$.
However, A* needs one traversal of edges to determine its heuristic upper bounds via coefficients $\lambda$ and $\delta$.
Given the range of these coefficients in Table~\ref{table:result_alg_leaf} (and Table~\ref{table:result_alg_ion} in Appendix), we can see that they are bounded to some deterministic values.
In particular, the heuristic factor $\delta$ is very close to one in all of the experiments (the minimum is 0.996), i.e., we practically have $\mathit{cost}_{\mathit{red}}\approx \mathit{cost}_{min}$.
For A* implemented with the reduced energy cost $\mathit{cost}_{\mathit{pot}}$, the range of the scaling factor $\lambda$ shows that the lower bound on the reduced cost can be as low as 84\% of the estimated energy when using the average energy efficiency over WLTP (given in Table~\ref{table_coefficient}).
Reducing the scaling factors to $\delta = 0.95$ and $\lambda = 0.80$ provides a safe margin that adequately scales the estimates, potentially removing the need for A* to tune its heuristic function before starting the search. The scaling factors can alternatively be calibrated once per map and per vehicle using a conservative scenario (e.g., no passenger load, corresponding to the lowest scaling factor) to ensure that the heuristic remains admissible under higher weight settings.

\subsection{Energy-Optimum Path Analysis}
Our next set of experiments investigates how vehicle dynamics affects energy-optimum paths. 
Here, we investigate the importance of vehicle dynamics in energy-efficient pathfinding for EVs by comparing the planned paths of different energy models.
For the experimental comparison, we used three different energy models, which take vehicle dynamics into account to varying degrees. 
The base case in this experiment is an energy model that uses road gradients and the EV average
energy efficiency, i.e., $\mathit{cost}_{b}=\overline{b_2}s^2+\overline{b_1}s+\overline{b}_0$. 
The second case adds mass as an extra parameter via $\mathit{cost}_{m}=(m\overline{a_2}+\overline{b_2})s^2+(m\overline{a_1}+\overline{b_1})s+(m\overline{a_0}+\overline{b_0})$. 
The third case incorporates driving patterns only as $\mathit{cost}_{d}={b_2}s^2+{b_1}s+{b}_0$ with coefficients now being link specific. 
The energy coefficients in these two cases are average among the patterns.
Finally, the last case considers our detailed model with both mass and driving patterns, as in Eq.~(\ref{eq:ener_eff}). The EV-based implementation of
the Bellman-Ford algorithm was used to find energy-optimum paths for the same graphs and node pairs as in our first experiment.

Table \ref{table_Avg_Dist} shows the result of incorporating vehicle dynamics into energy-optimal pathfinding. 
Extra mass (vehicle load) and driving pattern (speed profiles) are the parameters for which we study the length and the required energy of the resulting optimal paths. 
Column $\Delta p$ shows the percentage of paths that are different compared to when neither vehicle load nor speed profiles of roads is taken into account. Similarly, $\Delta l$ is the difference in path length, $\Delta \mathit{cost}$ is the difference in energy consumption and $\Delta \mathit{cost_{ef}}$ denotes the difference in the energy efficiency compared to the base case ($\mathit{cost_b}$).

\begingroup
\setlength{\tabcolsep}{3.5pt} 
\renewcommand{\arraystretch}{1.2} 
\begin{table*}[t]
\footnotesize
\centering
\caption{\small Path, distance and energy differences using vehicle dynamics parameters w.r.t a model that use road gradients only.}
\label{table_Avg_Dist}

\begin{tabular}{l| c| c | c |*{3}{c} |*{3}{c} |*{3}{c} |*{3}{c}}
\hline
 & Vehicle & Speed & $\Delta p$(\%) & \multicolumn{3}{c|}{ $\Delta l$(m)} & \multicolumn{3}{c|}{ $\Delta l$(\%)} &\multicolumn{3}{c|}{ $\Delta \mathit{cost}$(Wh)} & \multicolumn{3}{c}{$\Delta \mathit{cost_{ef}}$(Wh/100m)}\\ \cline{4-16}
City & load & profiles & Avg & Min & Avg & Max & Min & Avg & Max & Min & Avg & Max & Min & Avg & Max \\
 \hline

Munich & $\bullet$ & $\circ$ & 1.9 & 0 & 1 & 55 & 0.0 & 0.0 & 0.5 & -1 & 81 & 191 & -0.1 & 0.7 & 1.5\\
& $\circ$ & $\bullet$ & 18.1 & -18 & 47 & 1503 & -0.2 & 0.3 & 8.2 & 2 & 203 & 481 & 0.0 & 1.7 & 2.3 \\
$Avg |s|$: 0.9\% & $\bullet$ & $\bullet$ & 18.3 & -18 & 49 & 1503 & -0.2 & 0.3 & 8.2 & 16 & 316 & 725 & 1.0 & 2.7 & 3.3\\
 
 \hline
Milan & $\bullet$ & $\circ$ & 5.2 & 0 & 1 & 34 & 0.0 & 0.0 & 0.4 & 0 & 53 & 130 & 0.4 & 0.7 & 1.0\\
 & $\circ$ & $\bullet$ & 8.4 & 0  & 5 & 1277 & 0.0 & 0.1 & 11.1 & -2 & 141 & 309 &  -0.5 & 1.8 & 2.0\\
$Avg |s|$: 1.0\% & $\bullet$ & $\bullet$ & 10.0 & 0 & 7 & 1277 & 0.0 & 0.1 & 11.1 & 0 & 216 & 469 & 0.1 & 2.8 & 3.0\\
\hline

Calgary & $\bullet$ & $\circ$ & 6.6 & 0 & 1 & 127 & 0.0 & 0.0 & 1.2 & -30 & 142 & 417 & -0.5 & 0.7 & 2.1\\
& $\circ$ & $\bullet$ & 65.8 & -105 & 330 & 3407 & -0.6 & 1.4 & 11.4 & -150 & 121 & 471 & -1.3 & 0.6 & 2.5\\
$Avg |s|$: 1.6\% & $\bullet$ & $\bullet$ & 66.6 & -83 & 355 & 3388 & -0.5 & 1.5 & 11.4 & -22 & 300 & 795 & -0.6 & 1.5 & 3.7\\
 \hline

Canberra & $\bullet$ & $\circ$ & 4.0 & 0 & 1 & 146 & 0.0 & 0.0 & 1.4 & -5 & 119 & 304 & -0.4 & 0.7 & 2.0\\
 & $\circ$ & $\bullet$ & 55.8 & -82 & 224 & 2456 & -0.9 & 1.1 & 13.6 & -294 & 35 & 290 & -1.3 & 0.3 & 2.5\\
$Avg |s|$: 2.4\% & $\bullet$ & $\bullet$ & 55.9 & -82 & 240 & 2796 & -0.9 & 1.2 & 16.4 & -88 & 180 & 460 & -0.6 & 1.2 & 3.6\\
 \hline
 
Vancouver & $\bullet$ & $\circ$ & 10.4 & 0 & 2 & 141 & 0.0 & 0.0 & 2.0 & -22 & 49 & 146 & -1.1 & 0.8 & 3.2\\
 & $\circ$ & $\bullet$ & 12.1 & -75 & 4 & 302 & -1.0 & 0.0 & 4.8 & 5 & 121 & 312 & 0.5 & 1.9 & 3.1\\
$Avg |s|$: 2.5\% & $\bullet$ & $\bullet$ & 16.4 & -50 & 5 & 313 & -0.9 & 0.1 & 4.8 & 10 & 188 & 495 & 2.1 & 2.9 & 4.3\\
\hline

San & $\bullet$ & $\circ$ & 17.0 & 0 & 19 & 849 & 0.0 & 0.3 & 11.1 & -57 & 56 & 247 & -2.2 & 0.9 & 6.9\\
Francisco & $\circ$ & $\bullet$ & 16.6 & -899 & 37 & 1297 & -6.8 & 0.5 & 16.3 & 2 & 118 & 342 & -1.0 & 1.8 & 4.5\\
$Avg |s|$: 3.9\% & $\bullet$ & $\bullet$ & 22.3 & -90 & 52 & 1297 & -1.2 & 0.7 & 16.3 & 3 & 191 & 553 & 0.3 & 3.0 & 6.9\\
\hline
\end{tabular}
\end{table*}
\endgroup
\begin{figure*}[ht]
 \centering
 \footnotesize
 
  \begin{tabular}{cc|cc | cc}
   
     \hline
     & & &\\[-1.5\medskipamount]
     \label{fig:Calgary1}\includegraphics[trim=150 90 150 80, clip,width=0.25\columnwidth,scale=0.4]{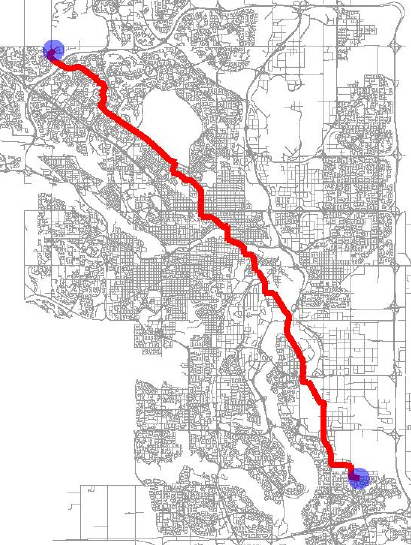}
     & \label{fig:Calgary2}\includegraphics[trim=150 90 150 80, clip,width=0.25\columnwidth,scale=0.4]{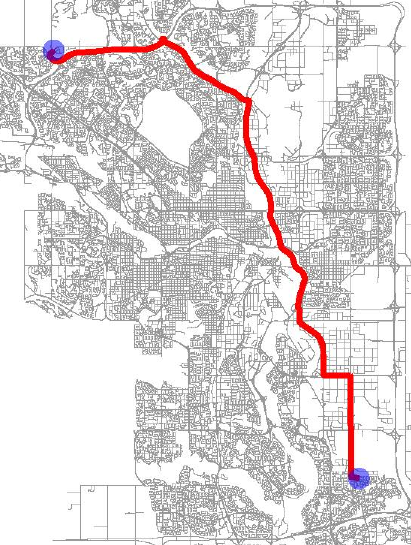}

    & \label{fig:Milan1}\includegraphics[trim=350 120 150 85, clip,width=0.25\columnwidth,scale=0.8]{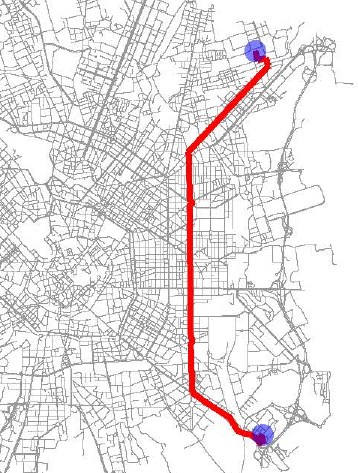}
     & \label{fig:Milan2}\includegraphics[trim=350 120 150 85, clip,width=0.25\columnwidth,scale=0.8]{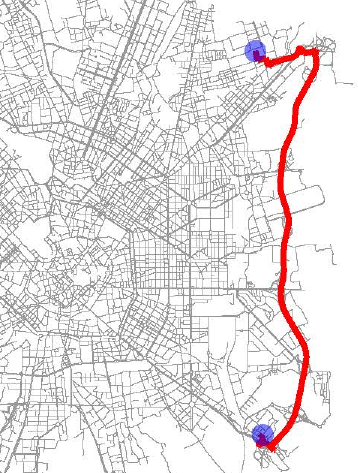} 
     
     & \label{fig:Munich1}\includegraphics[trim=220 60 330 80, clip,width=0.30\columnwidth,scale=1]{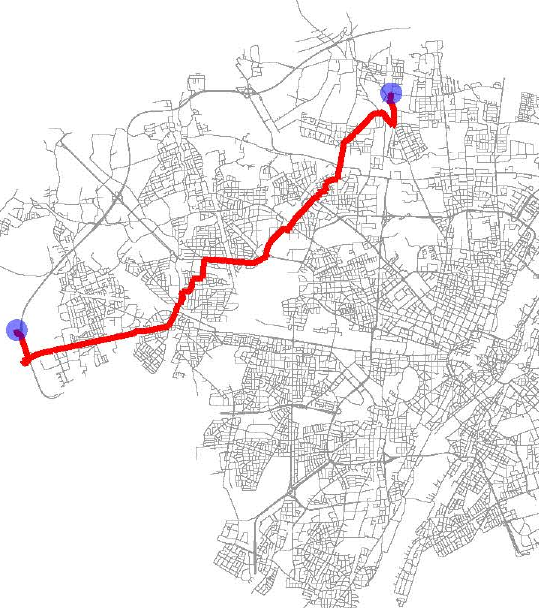}
    & \label{fig:Munich2}\includegraphics[trim=220 60 330 80, clip,width=0.30\columnwidth,scale=1]{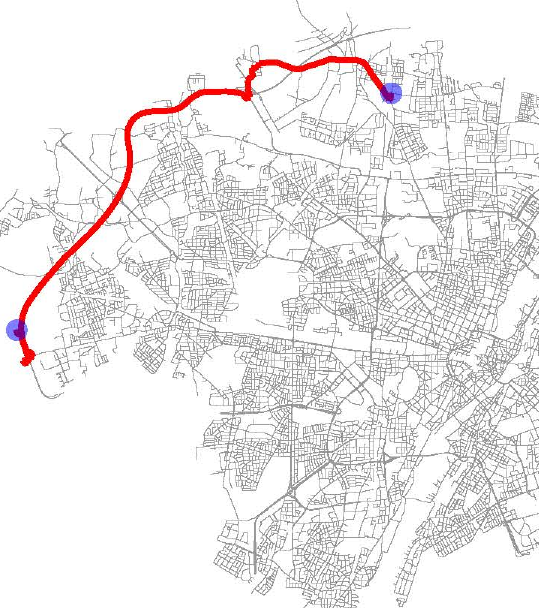}
    
     \\
     
     Calgary & $\Delta L= 2.7 km$ & Milan & $\Delta L= 1.3 km$ & Munich & $\Delta L= 1.5 km$\\ 
    

    
     
     \hline
     & & &\\[-1.5\medskipamount]
     \label{fig:SanFrans1}\includegraphics[trim=150 70 200 120, clip,width=0.25\columnwidth,scale=0.5]{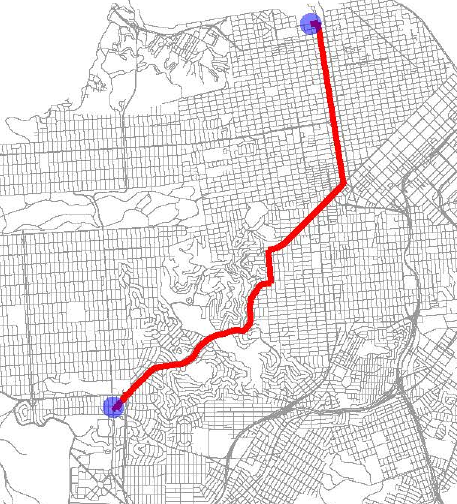}
     & \label{fig:SanFrans2}\includegraphics[trim=150 70 200 120, clip,width=0.25\columnwidth,scale=0.5]{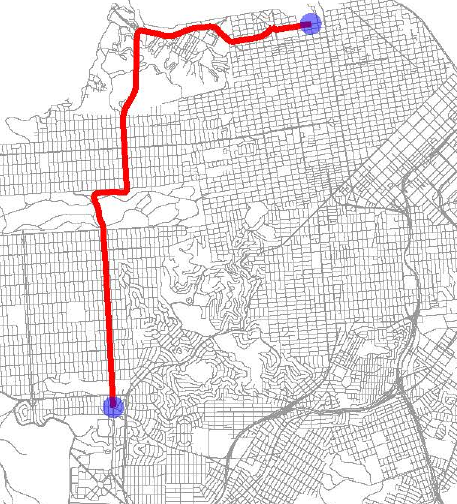}

     & \label{fig:Vancouver1}\includegraphics[trim=0 5 0 5, clip,width=0.25\columnwidth,scale=1]{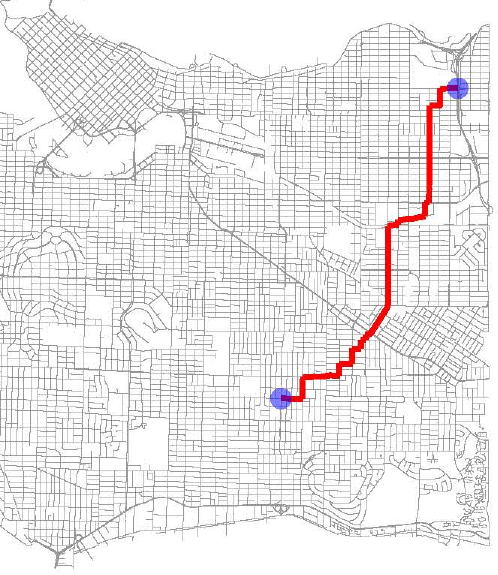}
     & \label{fig:Vancouver2}\includegraphics[trim=0 5 0 5, clip,width=0.25\columnwidth,scale=1]{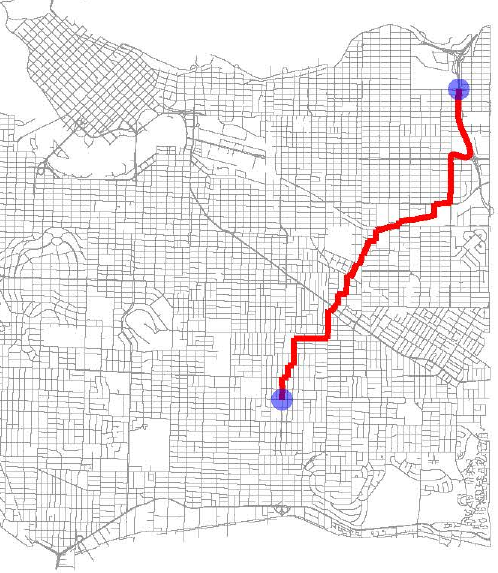}
     
     &  \label{fig:Canberra1}\includegraphics[trim=500 150 500 120, clip,width=0.24\columnwidth,scale=1]{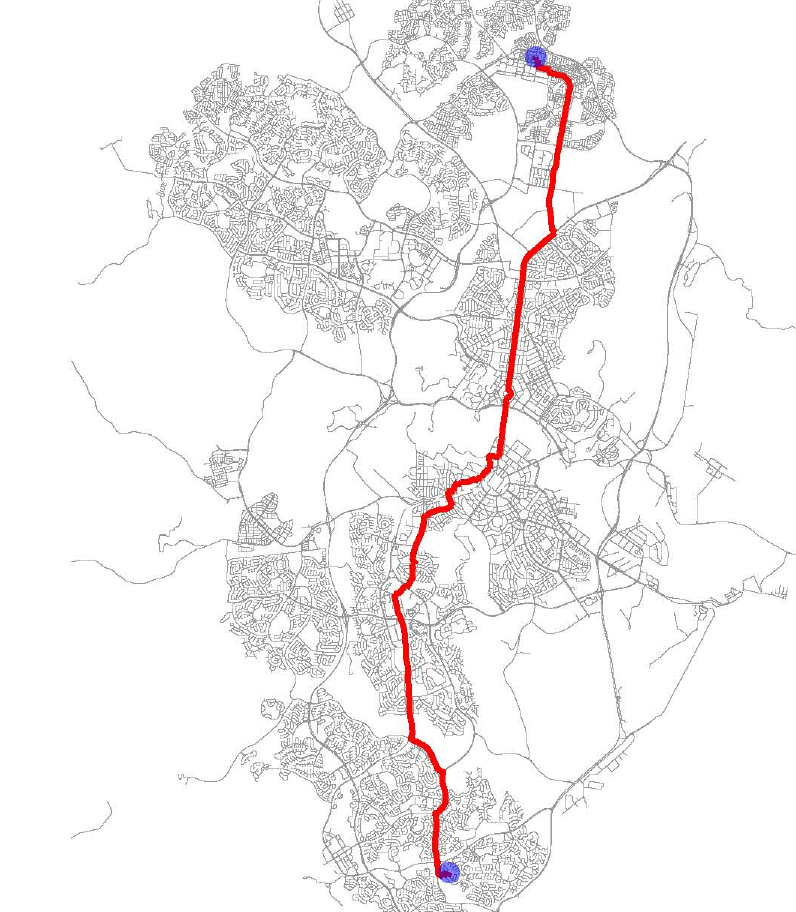}
    & \label{fig:Canberra2}\includegraphics[trim=500 150 500 120, clip,width=0.24\columnwidth,scale=1]{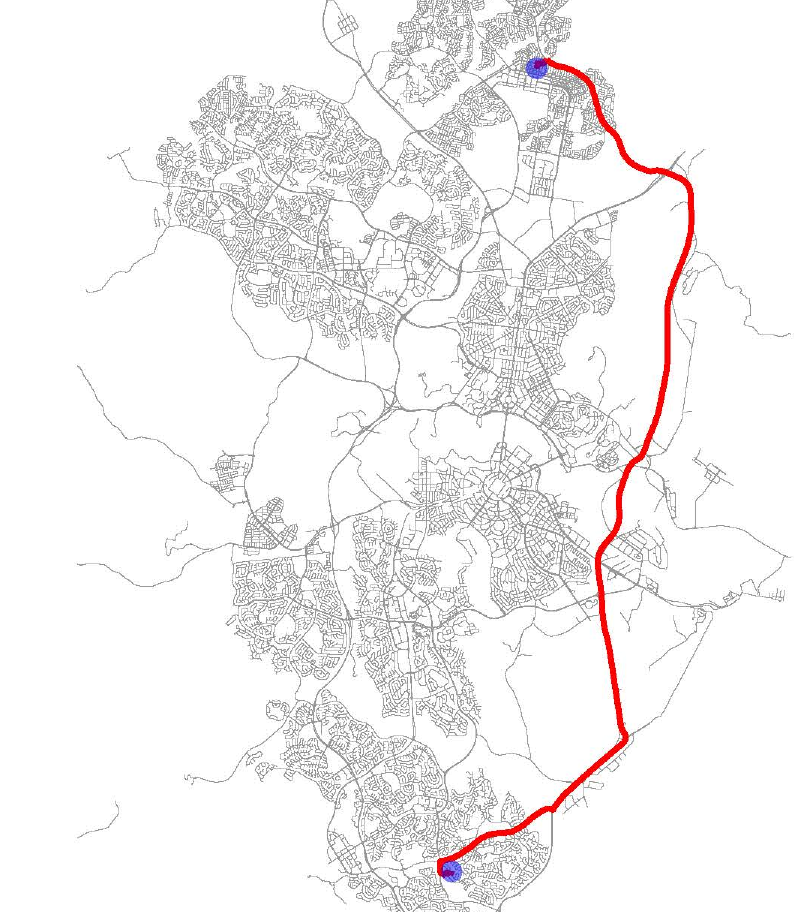}
    \\
     
     San Francisco & $\Delta L= 1.3 km$ & Vancouver & $\Delta L= 0.3 km$ & Canberra & $\Delta L= 2.8 km$\\ 
    
    \hline
     
  \end{tabular}
\caption{\small Planned routes for queries with maximum distance difference $\Delta l$ using a more accurate model. For each city, the left panel shows the route based only on road gradient, while the right panel shows the route computed using gradient, load, and driving pattern.}
 \label{routes_cities}
\end{figure*}
The results indicate that both vehicle mass and driving patterns can significantly affect energy estimates and, consequently, route planning.
When vehicle load is taken into account, the planned routes can differ by up to 17\%, although changes in path length remain limited (typically 1-2 meters across most maps). This is because vehicle mass has a linear impact on energy consumption, which generally does not alter the order of subpaths significantly.
In contrast, incorporating driving patterns leads to substantially greater differences, with up to 66\% variation in the traversed road segments (in the Calgary map). This is attributed to the distinct specification of driving patterns (energy coefficients), which introduces more variability in energy consumption across different road segments.
When both mass and driving patterns are considered together, the average path length can increase by up to 355 meters (in the Calgary map), and individual paths may show even more significant deviations.
Figure \ref{routes_cities} depicts two energy-optimum paths for a single \textit{(origin, destination)} pair in each city of this experiment, where the energy-optimum path using the more accurate energy model (right) is significantly longer than the one using the basic model (left), visualizing the maximum lengths difference provided in the fifth column of Table \ref{table_Avg_Dist}.
When considering changes in energy requirements, we find maximum differences of up to 469Wh compared to the base case (Milan map). 
Looking at the average energy consumption values, we observe that the base model underestimates energy requirements by up to 316Wh on average (Munich map).
Neglecting mass and driving patterns can also affect the estimated energy efficiency, where it can be miscalculated by up to 6.9Wh/100m (San Francisco map).
This amount of miscalculation can clearly lead to the resulting trip being infeasible, depending on initial battery levels. 
Our next experiment will study these effects further.

\subsection{Feasibility Study}
The third experiment of this study investigates the feasibility of routes executed without vehicle dynamics considerations. 
This experiment finds around 5,000 energy-optimum round-trips for random \textit{(origin, destination)} pairs in Calgary with the \textit{Peugeot iOn} as the test vehicle with 60\% initial SoC and four passengers on board. 
We first find the energy-optimum path in the road network with the basic energy model (without taking mass and driving pattern into account) and then recalculate the energy requirement of the same path with our accurate energy model in which vehicle dynamics is respected.

\begin{figure*}[t]
\footnotesize
\centering 
\includegraphics[width=1\textwidth]{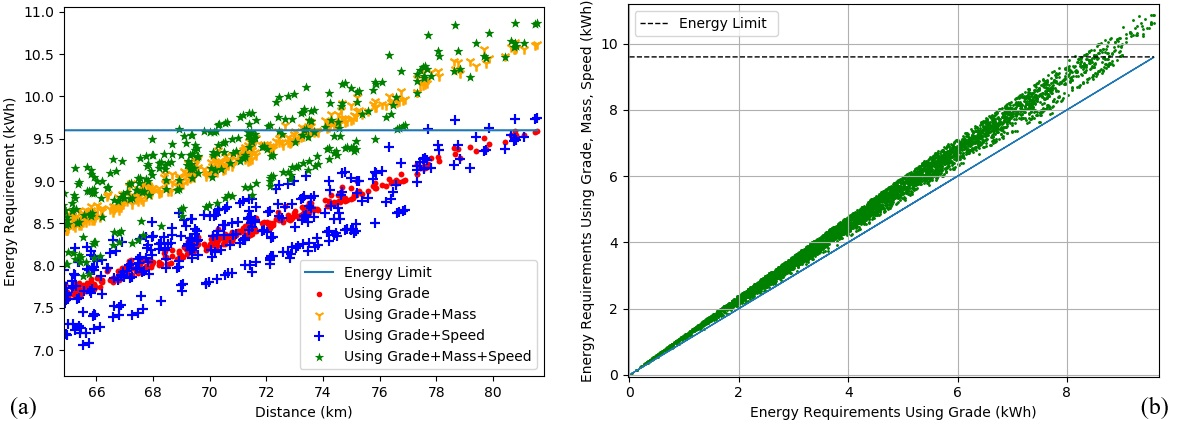} 
\caption{\small [Coloured] (a) Energy versus distance (enlarged for long trips), (b) Energy deviation.}
\label{feasibility1_calgary}
\end{figure*}

Figure \ref{feasibility1_calgary} focuses on the energy requirements of paths longer than 65km when different parameters of vehicle dynamics are considered. 
As we expected, adding mass to the energy model will increase the energy requirement of paths. 
Interestingly, incorporating the driving patterns may change the total energy cost in the opposite way, reducing the energy requirements on some trips. 
This reduction is primarily due to overestimating energy requirements on certain links along the path, where in reality, the speed profile demands less energy.
Meanwhile, it is less likely to get an overestimated path when both mass and driving patterns are considered. 
For the random trips shown in Figure \ref{feasibility1_calgary}, the first infeasible path is observed at a distance of 69km. 
Of all the trips that are longer than 69km (where the first infeasible trip is observed), 44\% of the routes would be infeasible if vehicle load and speed variation (driving patterns) are taken into account. 
Indeed, using these two parameters, all the trips with a distance of 76km and over are infeasible.

The graph in Figure \ref{feasibility1_calgary} (b) shows the error in energy calculation when our more accurate energy model (with vehicle load and driving pattern) is used. 
Analogously, trips that would consume more than initial energy (60\% SoC) are infeasible. 
This interpretation can easily be extended to other SoC percentages. 
Given any initial energy level in the range of battery capacity, there always exists a path for which its energy requirement is underestimated if vehicle load and driving pattern is to be ignored. 
One could consider a constant factor to compensate this energy miscalculation by scaling up the estimates, but this experiment shows that, in some cases, the percentage of energy deviation can be more than 10\%, a large factor when considering the full battery capacity (4kWh for a battery capacity of 40kWh).
Although it is true that infeasible routes may usually happen in low-battery cases, we believe that these are not corner cases. 
Most trips would be planned as return trips \textit{A-B-A}, or even \textit{multi-stop} trips, in which case even high initial SoCs may trigger infeasibility depending on the energy capacity, as we studied through this experiment for round-trips.
Hence, it is important to use algorithms that are:
i) equipped with energy models capable of accurately estimating energy costs; and
ii) able to rapidly replan in case of underestimation or changes in traffic conditions.
\subsection{Energy-time Trade-off}
A typical problem of energy-optimal path planning is that minor, slow roads appear to have the lowest energy consumption and the resulting paths, therefore, take much more time than the time-optimal path, rendering them unrealistic \cite{DBLP:journals/transci/BaumDGWZ19}. 
However, when taking realistic driving patterns into account, the picture changes. 
According to Table \ref{table_coefficient} where driving patterns are considered, the simulation results show that the slow roads will in fact have relatively higher energy consumption. 
This means, in general, slow pattern driving on a flat road would require more energy than driving with medium or fast driving patterns. 
This is mainly because of multiple stop-go actions, which reduce the overall system efficiency. 
Consequently, energy-optimum paths using our new, more realistic energy model does not necessarily prefer slow roads. 
However, our energy-optimal paths still require a (less pronounced) trade-off between energy and time. 
An interesting direction for future work is to consider efficient algorithms for improving this trade-off using our nonlinear energy model, setting an upper bound on the time/distance any path can take.

\section{Conclusion}
This paper presents a realistic energy model for EVs based on vehicle dynamics, which not only improves the solution quality of energy-optimal pathfinding but also enables the use of pre-processing-free pathfinding algorithms.
The results demonstrate that both vehicle load and driving patterns significantly influence energy estimates in path planning. 
Additionally, we have shown that for any initial energy level of an EV, the energy-optimal path may become infeasible if vehicle dynamics are not considered in the energy calculation.
From an algorithmic perspective, we identified a natural characteristic of the system model that allows for safe reweighting of the underlying graph in the presence of negative energy costs. 
Furthermore, we propose an alternative reweighting approach that leverages gravitational potential energy to obtain non-negative energy costs when detailed energy data is unavailable, as well as a heuristic guided search algorithm that further improves the average execution time and the number of expansions.
The nonlinear, more realistic energy model introduced in this paper can be broadly applied to standard pathfinding algorithms, using the proposed reweighting methods to achieve faster and more accurate energy-optimal pathfinding solutions.



%


\ifCLASSOPTIONcaptionsoff
  \newpage
\fi



\bibliographystyle{IEEEtran}
\bibliography{bibtex/bib/IEEEabrv,References}

\appendices
\section{Technical Proof}
\label{sec_appendix_A}
\noindent
\textit{Lemma 1. }
The function $\mathit{cost}-\mathit{cost_{pot}}$ always yields non-negative energy costs.

\begin{figure}
    \centering
    \footnotesize
    \tikzset{
  pics/.cd,
  car/.style = {
    code = {
    \shade[top color=white, bottom color=white, shading angle={135}]
        [draw=black,rounded corners=0.6ex,very thick] (0,.5) -- ++(0,1.3) -- ++(1,0) --  ++(2.8,0) -- ++(1.2,-0.2) -- ++(0,-1.1) -- (0,.5) -- cycle;
    \draw[very thick, rounded corners=0.5ex,fill=black!20!blue!20!white,thick, shading angle={135}]  (0.75,1.8) -- ++(0.6,0.7) -- ++(1.6,0) -- ++(0.8,-0.7) -- (0.75,1.8);
    
    \draw[semithick]  (1.25,1.5) -- (1.5,1.5);
    \draw[semithick]  (2.5,1.5) -- (2.75,1.5);
    
    \draw[thick]  (3.75,0.5) -- (3.75,1.8);
    \draw[thick]  (2.25,0.5) -- (2.25,2.5);
    \draw[thick]  (1.25,0.5) -- (0.75,1.8);
    
    \draw[draw=black,fill=gray!50,thick] (1,.5) circle (.5);
    \draw[draw=black,fill=gray!50,thick] (3.75,.5) circle (.5);
    \draw[draw=black,fill=white!80,semithick] (1,.5) circle (.35);
    \draw[draw=black,fill=white!80,semithick] (3.75,.5) circle (.35);
    \draw[draw=black,fill=gray!80,semithick] (1,.5) circle (.1);
    \draw[draw=black,fill=gray!80,semithick] (3.75,.5) circle (.1);

    }
  },
  car 2/.style = {
  \draw[thick]  (3.75,0.5) -- (3.75,1.8);
  }
}
\begin{tikzpicture}[scale=0.6,every node/.style={scale=0.25}]
\draw[rounded corners=1.0ex,very thick,black] (0,0) -- ++(3,0)-- ++(3,2)-- ++(3,0) -- ++(2,-1) -- ++ (3,0);
\draw[dashed,thick,black] (2.5,0) -- (14,0);
\draw[dashed,thick,black] (8.75,1) -- (11,1);
\draw[->,thick,black]  (6.25,0.1) -- node [right] {\Huge $\Delta h_1$}  (6.25,1.9);
\draw[<-,thick,black]  (8.75,1.1) -- node [right] {\Huge $\Delta h_2$}  (8.75,1.9);
\draw[->,thick,black]  (14,0.1) -- node [right] {\Huge $\Delta H$}  (14,0.9);
\draw[->,semithick,black]  (3.5,0.75) -- node [sloped,above] {\Huge 1} (5,1.70) ; %
\draw[->,semithick,black]  (9.75,2) -- node [sloped,above] {\Huge 2} (10.75,1.5);
\path (0,0) pic {car} (6.5,2) pic [opacity=0.2]{car} (12,1) pic [opacity=0.2]{car};

\end{tikzpicture}
    \caption{A sample scenario for an EV traversing links with different gradients where $\Delta H=\Delta h_1+\Delta h_2$.}
    \label{fig:potential_energy}
\end{figure}
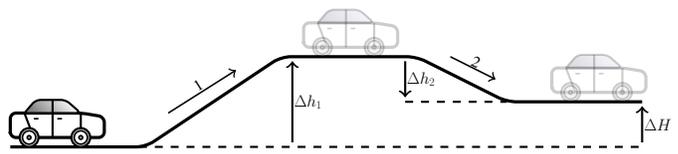

\begin{proof}
EVs may exhibit negative energy costs only on negative slopes, while the energy requirement on positive gradients is always positive. Note that a negative gradient link may still have a positive energy cost, e.g.,\ due to friction or acceleration.

First, consider link 1 in Figure~\ref{fig:potential_energy} with a positive slope. 
If the EV traverses this link, its relative height will increase by $\Delta h_1$. 
In other words, the EV will have a greater gravitational potential energy at the end node of the link.
Based on the law of conservation of energy, the change in one form of energy (potential energy here) is compensated by the other form (electrical energy).
Therefore, the EV must at least consume $\mathit{cost_{pot}}$ to compensate the gravitational energy over uphill links, i.e., $ \mathit{cost} \ge \mathit{cost_{pot}}$ or $\mathit{cost}-\mathit{cost_{pot}} \ge 0$. Therefore, the reduced cost is non-negative for the uphill edges. 

Second, consider links with negative ground slopes, for example, link 2 in Figure~\ref{fig:potential_energy}.
For this link, since the relational change in the height ($\Delta h_2$) of the EV decreases, the gravitational potential energy is thus negative or $\mathit{cost_{pot}}<0$.
Let us assume the most optimistic scenario where the EV does not need to consume any kind of electrical energy for propulsion and has enough capacity to store the recuperated energy.
In this condition, since the energy is transformed from the potential to electrical form, the maximum amount of energy the EV can regenerate by gliding on a negative slope is again limited by the gravitational potential energy, i.e. $ \mathit{cost_{pot}} \le \mathit{cost}$ or $\mathit{cost}-\mathit{cost_{pot}} \ge 0$.
Therefore, the reduced cost is also non-negative for the downhill edges.

Third, let us assume that $\mathit{cost}$ of a link needs to be adjusted to $\mathit{cost}'$ to meet the battery limits.
Since we never reduce $\mathit{cost}$ via the adjustment, we have $\mathit{cost} \le \mathit{cost}'$ which yields $\mathit{cost}'-\mathit{cost_{pot}} \ge 0$ based on the both cases above.
\end{proof}

\textit{Lemma 2. }
The energy heuristic $h_e=\lambda \eta_e  h_d$ is consistent and admissible.

\begin{proof}
Based on the natural characteristics of roads network where triangle inequality holds for distances (e.g., euclidean or spherical distance), denoting $h_d$ as a consistent distance heuristic, we have
\begin{equation} \label{eq:ener_pot_cons01}
h_d(u) \le l_{uv}+h_d(v) \quad or \quad h_d(u) - h_d(v) \le l_{uv} 
\end{equation}
Since both $\lambda$ and $\eta_e $ are non-negative, we can extend Eq.~(\ref{eq:ener_pot_cons01}) to
\begin{equation} \label{ener_pot_cons02}
\lambda \eta_e (h_d(u) - h_d(v)) \le \lambda \eta_e l_{uv}
\end{equation}
According to Eq.~(\ref{eq:ener_heur_pot}), Eq.~(\ref{ener_pot_cons02}) is the equivalent of
\begin{equation} \label{ener_pot_cons03}
h_e(u) - h_e(v) \le \lambda \eta_e l_{uv}
\end{equation}
Since the value of $\lambda$ is limited by $\mathit{cost}_{\mathit{red}}(u,v)/(\eta_e l_{uv})$ by the definition, we also have
\begin{equation} \label{ener_pot_cons04}
\lambda \eta_e l_{uv} \le cost_{red}(u,v)
\end{equation}
which finally yields
\begin{equation} \label{ener_pot_cons05}
h_e(u) - h_e(v) \le cost_{red}(u,v)
\end{equation}
Therefore, the given energy heuristic $h_e$ is consistent.
Furthermore, we already know that every consistent heuristic function $h$ with $h(\mathit{goal}) = 0$ is admissible \cite{pearl1984heuristics}.
Thus, given $h_d(\mathit{goal})=h_e(goal)=0$, the energy heuristic $h_e$ is admissible.

The proposed energy heuristic is consistent and admissible
even with energy limit considerations.
In special cases, negative energies cannot be fully captured by the EV's battery and therefore the energy adjustment is necessary.
This adjustment does not affect our energy heuristic since the resulting energy cost is always larger than the original energy and adjustments cannot decrease the links' energy requirement.
For example, assume a situation where a fully charged EV faces a link with the energy requirement $\mathit{cost}_e < 0$. Since the EV cannot store any portion of the link's energy (its charge level is full), the energy requirement of the link is then adjusted to $\mathit{cost}'_e=0$. Therefore, the adjusted energy is obviously larger than $\mathit{cost}_e$, i.e., $\mathit{cost}_e \le \mathit{cost}'_e$, which is in favour of $\mathit{cost}_{\mathit{red}}$.
\end{proof}

\textit{Lemma 3. }
The heuristic $h_e=\delta (m{a_0}_{min}+{b_0}_{min})h_d$ is consistent and admissible.

\begin{proof}
 Based on the natural characteristics of roads network where triangle inequality holds for distances and given $h_d$ as a consistent distance heuristic, we have
\begin{equation} \label{eq:ener_pot_cons1}
h_d(u) \le l_{uv}+h_d(v) \quad or \quad h_d(u) - h_d(v) \le l_{uv} 
\end{equation}
Since both ${a_3}_{min}$ and ${b_3}_{min}$ are positive and also $0\le \delta $, we can extend Eq.~(\ref{eq:ener_pot_cons1}) to
\begin{equation} \label{ener_pot_cons2}
\delta(m{a_3}_{min}+{b_3}_{min})(h_d(u) - h_d(v)) \le \delta(m{a_3}_{min}+{b_3}_{min})l_{uv}
\end{equation}
According to Eq.~(\ref{eq:ener_heur_pi},~\ref{eq:ener_min}), Eq.~(\ref{ener_pot_cons2}) is the equivalent of
\begin{equation} \label{ener_pot_cons3}
h_e(u) - h_e(v) \le \delta cost_{min}(u,v)
\end{equation}
Since the value of $\delta$ is limited by $\mathit{cost}_{\mathit{red}}(u,v)/ \mathit{cost}_{min}(u,v)$, we also have
\begin{equation} \label{ener_pot_cons4}
\delta cost_{min}(u,v) \le cost_{red}(u,v)
\end{equation}
which finally yields
\begin{equation} \label{ener_pot_cons5}
h_e(u) - h_e(v) \le cost_{red}(u,v)
\end{equation}
Therefore, the given energy heuristic $h_e$ is consistent.
Furthermore, we already know that every consistent heuristic function $h$ with $h(\mathit{goal}) = 0$ is admissible \cite{pearl1984heuristics}.
Thus, given $h_d(\mathit{goal})=h_e(goal)=0$, the energy heuristic $h_e$ is admissible.

Similar to our previous (model-independent) heuristic function, this energy heuristic is consistent and admissible
even with energy limit considerations as that would always increase $\mathit{cost}_{\mathit{red}}$.
\end{proof}

\textit{Theorem 1.}
The energy-optimal A* search (Algorithm~\ref{alg:Dij}) returns the minimum energy requirement among all $u_o$--$u_d$ paths.

\begin{proof}
Under the premises of Lemma~1, the reduced cost $\mathcal{C}$ of all paths is non-negative because the reduced edge cost $\mathit{cost}_{\mathit{red}}$ has already been shown to be non-negative.
Lemmas~2 and~3 further establish that the heuristic function $h_e$ is consistent and admissible with respect to the reduced cost $\mathit{cost}_{\mathit{red}}$.
Consequently, vertices in the A* search are expanded in non-decreasing order of their $f$-values, and $\mathcal{C}(u)$ represents the lowest reduced cost required to reach state $u$ from $\mathit{start}$ when $u$ is expanded by A* \cite{hart1968formal}.
Furthermore, Algorithm~\ref{alg:Dij} always uses the actual energy requirement of partial paths to enforce battery constraints (lines~\ref{alg2:energy-check1}-\ref{alg2:energy-check2}), ensuring that energy limits are respected correctly throughout the search.
Therefore, when the search terminates, either the destination $u_d$ has been reached with the smallest reduced cost $\mathcal{C}(u_d)$, or the open list $\mathcal{Q}$ becomes empty.
The latter case implies that no feasible solution exists and the algorithm correctly returns $\infty$.
In the former case, a feasible path exists.
Since all $u_o$--$u_d$ paths share the same $\mathit{cost}_{\mathit{pi}}$ and $\mathit{cost}_{\mathit{pot}}$ (as these height-induced energy components are path-independent), and because the available energy at each visited vertex $u$ along optimal subpaths is tracked by $\mathcal{E}(u)$, the value $\mathcal{E}(u_d)$ denotes the maximum possible remaining energy at the destination when following the least reduced-cost path.
Therefore, the required initial energy can be computed as $\mathcal{E}_{\mathit{init}} - \mathcal{E}(u_d)$.
Thus, the energy-optimal A* algorithm correctly returns the minimum energy requirement among all feasible paths between $u_o$ and $u_d$.
\end{proof}

\section{Extended Experiments}
\label{sec_appendix_B}
Tables below present statistical results on runtime and parameters for tow other EVs considered in this study (same test cases).

\renewcommand{\arraystretch}{1.1} 
\begin{table}[ht]
\centering
\footnotesize
\caption{\small Experiment results and parameters for \textit{Peugeot iOn} with no passenger. The last column shows average number of expansions.}
\setlength{\tabcolsep}{3.0pt}
\begin{tabular}{l|l|l | r| *{2}{r} | r}
\hline
City & Algorithm & Algorithm &\multicolumn{3}{c|}{ Runtime(s)} & \multicolumn{1}{c}{{Exp.}}\\ \cline{4-6} 
Network & Name & Parameters &  Prep.& Avg & Max & $\times10^3$\\
\hline

Munich & Bellman-Ford &  & 0.000 & 12.59 & 22.59 & 86.0\\
Nodes: & Johnson &   & 1.435 & 0.375 & 0.738 & 7.8\\ 
13,974 & Johnson-$\pi_h$ & $\alpha \in$[139,387]  & 0.416 & 0.369 & 0.738 & 7.8\\
Arcs: & Dijkstra-$\mathit{cost_{pot}}$ & $mg+Mg$ = 287 &  0.000 & 0.360 & 0.725 & 7.8\\
36,228 & Dijkstra-$\mathit{cost_{pi}}$ & $m\overline{a}+\overline{b}$ = 270 & 0.000 & 0.359 & 0.738 & 7.8\\
 & {A*-$\mathit{cost}_{pot}$} &  $\lambda$ = 0.841  & 0.414 &  0.179 & 0.654 & 3.4\\
  & {A*-$\mathit{cost}_{pi}$} & $\delta$ = 0.996 &  0.416 & 0.171 & 0.650 & 3.2\\
\hline

Milan & Bellman-Ford &  &  0.000 & 8.546 & 14.74  & 69.9\\
Nodes: & Johnson &  &  1.104 & 0.303 & 0.592 & 7.1\\ 
13,377 & Johnson-$\pi_h$ & $\alpha \in$[138, 391] &  0.303 & 0.276 & 0.574 & 7.1\\
Arcs: & Dijkstra-$\mathit{cost_{pot}}$ & ${mg+Mg}$ = 287 &  0.000 & 0.270 & 0.550 & 7.1\\
26,539 & Dijkstra-$\mathit{cost_{pi}}$ & $m\overline{a}+\overline{b}$ = 270 &  0.000 & 0.269 & 0.547 & 7.1\\
  & {A*-$\mathit{cost}_{pot}$} &  $\lambda$ = 0.841 & 0.304 &  0.129 & 0.431 & 2.9\\
 & {A*-$\mathit{cost}_{pi}$} & $\delta$ = 0.996  & 0.302 &  0.123 & 0.439 & 2.7\\
\hline

Calgary & Bellman-Ford &  &  0.000 & 32.43 & 61.06  & 242.5\\
Nodes: & Johnson &  &  3.794  & 0.821 & 1.927& 17.1\\ 
32,603  & Johnson-$\pi_h$ & $\alpha \in$[139, 389] & 1.097 & 0.796 & 1.869 & 17.0\\
Arcs: & Dijkstra-$\mathit{cost_{pot}}$ & $mg+Mg$ = 287 &  0.000 & 0.782 & 1.851 & 17.0\\
77,172 & Dijkstra-$\mathit{cost_{pi}}$ & $m\overline{a}+\overline{b}$ = 270 &  0.000 & 0.783 & 1.882 & 17.0\\
 & {A*-$\mathit{cost}_{pot}$} &  $\lambda$ = 0.840  & 1.101 &  0.389 & 1.680 & 7.2\\
 & {A*-$\mathit{cost}_{pi}$} & $\delta$ = 0.996  & 1.099 &  0.369 & 1.481 & 6.8\\
\hline

Canberra & Bellman-Ford &  & 0.000 & 16.92 & 30.09 & 149.4\\
Nodes: & Johnson &  & 2.486 & 0.461 & 0.991 & 11.4\\ 
23,594 & Johnson-$\pi_h$ & $\alpha \in$[139, 391] &  0.595 & 0.453 & 1.000 & 11.4\\
Arcs: & Dijkstra-$\mathit{cost_{pot}}$ & $mg+Mg$ = 287 &  0.000 & 0.438 & 0.991 & 11.4\\
52,112 & Dijkstra-$\mathit{cost_{pi}}$ & $m\overline{a}+\overline{b}$ = 270 &  0.000 & 0.438 & 0.941 & 11.4\\
  & {A*-$\mathit{cost}_{pot}$} &  $\lambda$ = 0.840 &  0.596 &  0.224 & 0.984 & 5.0\\
 & {A*-$\mathit{cost}_{pi}$} & $\delta$ = 0.996 &  0.597 & 0.213 & 0.922 & 4.8\\
\hline

Vancouver & Bellman-Ford &  &  0.000 & 4.020 & 6.945  & 21.2\\
Nodes: & Johnson &   &  1.033 & 0.191 & 0.434& 3.7\\ 
7,599 & Johnson-$\pi_h$ & $\alpha \in$[138, 391] &  0.269 & 0.188 & 0.430 & 3.7\\
Arcs: & Dijkstra-$\mathit{cost_{pot}}$ & $mg+Mg$ = 287 &  0.000 & 0.184 & 0.427  & 3.7\\
23,012  & Dijkstra-$\mathit{cost_{pi}}$ & $m\overline{a}+\overline{b}$ = 270 & 0.000 & 0.184 & 0.434 & 3.7\\
  & {A*-$\mathit{cost}_{pot}$} &  $\lambda$ = 0.840 &  0.270 &  0.086 & 0.419 & 1.5\\
 & {A*-$\mathit{cost}_{pi}$} & $\delta$ = 0.996 & 0.268 & 0.081 & 0.416 & 1.4\\

\hline

San Fran.  & Bellman-Ford &  &  0.000 & 6.272 & 14.48  & 37.6\\
Nodes: & Johnson &  &  2.545 & 0.223 & 0.516 & 4.5\\ 
9,555 & Johnson-$\pi_h$ & $\alpha \in$[139, 391]  &  0.312 & 0.216 & 0.503 & 4.5\\
Arcs: & Dijkstra-$\mathit{cost_{pot}}$ & $mg+Mg$ = 287 &  0.000 & 0.210 & 0.482 & 4.5\\
26,809  & Dijkstra-$\mathit{cost_{pi}}$ & $m\overline{a}+\overline{b}$ = 270 &  0.000 & 0.210 & 0.484 & 4.5\\
 & {A*-$\mathit{cost}_{pot}$} &  $\lambda$ = 0.840 & 0.314 &  0.097 & 0.459 & 1.8\\
 & {A*-$\mathit{cost}_{pi}$} & $\delta$ = 0.996 &  0.312 & 0.091 & 0.456 & 1.7\\
\hline
\end{tabular}
\label{table:result_alg_ion}
\end{table}

\renewcommand{\arraystretch}{1.1} 
\begin{table}[ht]
\centering
\footnotesize
\caption{\small Experiment results and parameters for the \textit{GM-EV1} with only one passenger. The last column shows average number of expansions.}
\setlength{\tabcolsep}{3.0pt}
\begin{tabular}{l|l|l | r| *{2}{r} | r}
\hline
City & Algorithm & Algorithm &\multicolumn{3}{c|}{ Runtime(s)} & \multicolumn{1}{c}{{Exp.}}\\ \cline{4-6} 
Network & Name & Parameters &  Prep.& Avg & Max & $\times10^3$\\
\hline

Munich & Bellman-Ford &  & 0.000 & 13.12 & 22.77  & 85.7\\
Nodes: & Johnson &   & 2.137  & 0.421 & 0.955 & 7.8\\ 
13,974 & Johnson-$\pi_h$ & $\alpha \in$[286,531]  & 0.628 & 0.414 & 0.945 & 7.8\\
Arcs: & Dijkstra-$\mathit{cost_{pot}}$ & $mg+Mg$ = 416 &  0.000 & 0.404 & 0.857 & 7.8\\
36,228 & Dijkstra-$\mathit{cost_{pi}}$ & $m\overline{a}+\overline{b}$ = 418 & 0.000 & 0.405 & 0.873 & 7.8\\
  & A*-$\mathit{cost_{pot}}$ &  $\lambda$ = 0.930 & 0.626 & 0.180 & 0.745 & 3.0\\
 & A*-$\mathit{cost_{pi}}$ & $\delta$ = 0.999 & 0.625 &  0.180 & 0.731 & 3.0 \\
\hline

Milan & Bellman-Ford &  &  0.000 & 8.693 & 15.07  & 68.4\\
Nodes: & Johnson &  &  1.672  & 0.326 & 0.782 & 7.1\\ 
13,377 & Johnson-$\pi_h$ & $\alpha \in$[278, 540] &  0.450 & 0.319 & 0.724 & 7.1\\
Arcs: & Dijkstra-$\mathit{cost_{pot}}$ & ${mg+Mg}$ = 416 &  0.000 & 0.310 & 0.760 & 7.1\\
26,539 & Dijkstra-$\mathit{cost_{pi}}$ & $m\overline{a}+\overline{b}$ = 418 &  0.000 & 0.310 & 0.732 & 7.1\\
  & A*-$\mathit{cost_{pot}}$ &  $\lambda$ = 0.930 & 0.452 & 0.131 & 0.491 & 2.5\\
 & A*-$\mathit{cost_{pi}}$ & $\delta$ = 0.999 & 0.453 & 0.131 & 0.492 & 2.5\\
\hline

Calgary & Bellman-Ford &  &  0.000 & 27.94 & 50.27  & 241.5\\
Nodes: & Johnson &  &  3.860  & 0.738 & 1.662 & 17.1\\ 
32,603  & Johnson-$\pi_h$ & $\alpha \in$[289, 532] & 0.901 & 0.716 & 1.509 & 17.0\\
Arcs: & Dijkstra-$\mathit{cost_{pot}}$ & $mg+Mg$ = 416 &  0.000 & 0.701 & 1.513 & 17.0\\
77,172 & Dijkstra-$\mathit{cost_{pi}}$ & $m\overline{a}+\overline{b}$ = 418 &  0.000 & 0.700 & 1.466 & 17.0\\
 & A*-$\mathit{cost_{pot}}$ &  $\lambda$ = 0.930 & 0.900 &  0.312 & 1.253 & 6.4 \\
 & A*-$\mathit{cost_{pi}}$ & $\delta$ = 0.999 & 0.897 & 0.312 & 1.236 & 6.5 \\
\hline

Canberra & Bellman-Ford &  & 0.000 & 16.97 & 32.24 & 148.1\\
Nodes: & Johnson &  & 4.063 & 0.471 & 1.175  & 11.3\\ 
23,594 & Johnson-$\pi_h$ & $\alpha \in$[286, 538] &  0.663 & 0.464 & 1.091 & 11.4\\
Arcs: & Dijkstra-$\mathit{cost_{pot}}$ & $mg+Mg$ = 416 &  0.000 & 0.452 & 1.035 & 11.4\\
52,112 & Dijkstra-$\mathit{cost_{pi}}$ & $m\overline{a}+\overline{b}$ = 418 &  0.000 & 0.452 & 1.049 & 11.4\\
  & A*-$\mathit{cost_{pot}}$ &  $\lambda$ = 0.930 &  0.665 & 0.209 & 0.907 & 4.5\\
 & A*-$\mathit{cost_{pi}}$ & $\delta$ = 0.999 & 0.661  & 0.209 & 0.950 & 4.6\\
\hline

Vancouver & Bellman-Ford &  &  0.000 & 3.930 & 7.642 & 20.9\\
Nodes: & Johnson &   &  1.925 & 0.247 & 0.567 & 3.7\\ 
7,599 & Johnson-$\pi_h$ & $\alpha \in$[284, 552] &  0.353 & 0.241 & 0.572 & 3.7\\
Arcs: & Dijkstra-$\mathit{cost_{pot}}$ & $mg+Mg$ = 416 &  0.000 & 0.236 & 0.546 & 3.7\\
23,012  & Dijkstra-$\mathit{cost_{pi}}$ & $m\overline{a}+\overline{b}$ = 418 & 0.000 & 0.236 & 0.548 & 3.7\\
  & A*-$\mathit{cost_{pot}}$ &  $\lambda$ = 0.930 & 0.352 & 0.101 & 0.568 & 1.3 \\
 & A*-$\mathit{cost_{pi}}$ & $\delta$ = 0.999 & 0.349   & 0.101 & 0.573 & 1.3 \\
\hline

San Fran.  & Bellman-Ford &  &  0.000 & 6.343 & 16.66 & 37.2\\
Nodes: & Johnson &  &  5.563 & 0.276 & 0.684 & 4.6\\ 
9,555 & Johnson-$\pi_h$ & $\alpha \in$[284, 541]  &  0.435 & 0.264 & 0.660 & 4.5\\
Arcs: & Dijkstra-$\mathit{cost_{pot}}$ & $mg+Mg$ = 416 &  0.000 & 0.258 & 0.653 & 4.5\\
26,809  & Dijkstra-$\mathit{cost_{pi}}$ & $m\overline{a}+\overline{b}$ = 418 &  0.000 & 0.258 & 0.645 & 4.5\\
  & A*-$\mathit{cost_{pot}}$ &  $\lambda$ = 0.930 & 0.433 & 0.110 & 0.596 & 1.6\\
& A*-$\mathit{cost_{pi}}$ & $\delta$ = 0.999 & 0.431  & 0.110 & 0.600 & 1.6\\
\hline
\end{tabular}
\label{table:result_alg_GM}
\end{table}

\end{document}